\documentclass{article} 
\usepackage{iclr2025_conference,times}

\usepackage{amsmath,amsfonts,bm}

\def\eqref#1{equation~\ref{#1}}

\def\1{\bm{1}}

\DeclareMathAlphabet{\mathsfit}{\encodingdefault}{\sfdefault}{m}{sl}
\SetMathAlphabet{\mathsfit}{bold}{\encodingdefault}{\sfdefault}{bx}{n}

\usepackage{hyperref}
\usepackage{url}
\usepackage{multicol}
\usepackage{multirow}
\usepackage{subfigure}
\usepackage{comment}
\usepackage{graphicx} 
\usepackage{booktabs} 
\usepackage{wrapfig}

\title{CrossMPT: Cross-attention Message-Passing Transformer for Error Correcting Codes}

\author{Seong-Joon Park\\
POSTECH\\
\texttt{seongjoon@postech.ac.kr}\\
\And
Hee-Youl Kwak\\
University of Ulsan\\
\texttt{hykwak@ulsan.ac.kr}\\
\And
Sang-Hyo Kim\\
Sungkyunkwan University\\
\texttt{iamshkim@skku.edu}\\
\And
Yongjune Kim\\
POSTECH\\
\texttt{yongjune@postech.ac.kr}\\
\And
Jong-Seon No\\
Seoul National University\\
\texttt{jsno@snu.ac.kr}
}

\iclrfinalcopy 
\begin{document}

\maketitle

\begin{abstract}
Error correcting codes~(ECCs) are indispensable for reliable transmission in communication systems.~The recent advancements in deep learning have catalyzed the exploration of ECC decoders based on neural networks.~Among these, transformer-based neural decoders have achieved state-of-the-art decoding performance.
In this paper, we propose a novel Cross-attention Message-Passing Transformer~(CrossMPT), which shares key operational principles with conventional message-passing decoders.
While conventional transformer-based decoders employ self-attention mechanism without distinguishing between the types of input vectors (i.e., magnitude and syndrome vectors), CrossMPT updates the two types of input vectors separately and iteratively using two masked cross-attention blocks.~The mask matrices are determined by the code's parity-check matrix, which explicitly captures the irrelevant relationship between two input vectors.
Our experimental results show that CrossMPT significantly outperforms existing neural network-based decoders for various code classes.
Notably, CrossMPT achieves this decoding performance improvement, while significantly reducing the memory usage, complexity, inference time, and training time.

\end{abstract}

\section{Introduction}
\label{sec_intro}
The fundamental objective of digital communication systems is to reliably transmit information from source to destination through noisy channels.
Error correcting codes~(ECCs) are crucial for ensuring the integrity of transmitted data in digital communication systems.
The advancements in deep learning across diverse tasks, such as natural language processing~(NLP), image classification, or object detection~\citep{b_bert,b_resnet,b_od1,b_detr}, have motivated the application of deep learning techniques to ECC decoders.
This has led to the development of neural decoders~\citep{b_Kim2018,b_Kim2020,b_Nachmani2016,b_Nachmani2018,b_Dai2021,b_Lugosch2017}.
The key aim of these neural decoders is to improve decoding performance by overcoming limitations of the conventional decoders such as belief propagation~(BP)~\citep{b_BP} or min-sum~(MS)~\citep{b_MS} decoders.

Among neural decoders, model-free neural decoders employ an arbitrary neural network architecture~(e.g., deep neural networks~\citep{b_Gruber2017}, recurrent neural networks~\citep{b_Bennantan2018} and transformers~\citep{b_ECCT,b_DDECC,b_Park2023,b_Choukroun2024,b_Choukroun2024_ICML}) as the ECC decoder, without relying on prior knowledge of specific decoding algorithms.
Since model-free neural decoders are not based on specific decoding algorithms, their training is prone to overfitting, largely due to the exponentially large number of codewords~\citep{b_Bennantan2018}.
To circumvent overfitting, these neural decoders incorporate a preprocessing step where the magnitude and syndrome vectors from the received codeword are concatenated and used as inputs. 
The preprocessing step is essential for integrating an effective network architecture for the ECC decoder without an overfitting issue~\citep {b_Bennantan2018}. 
For example, transformer-based ECC decoders~\citep{b_ECCT,b_DDECC,b_Park2023,b_Choukroun2024,b_Choukroun2024_ICML} achieve state-of-the-art decoding performance.
However, two important questions have not been addressed: 1) how to effectively manage the two distinct input vectors (magnitude and syndrome), and 2) how to design an efficient transformer-based decoder architecture. 



Conventional transformer-based ECC decoders, initially proposed as Error Correction Code Transformer~(ECCT)~\citep{b_ECCT,b_DDECC,b_Park2023,b_Choukroun2024,b_Choukroun2024_ICML}, receive the concatenated magnitude and syndrome embeddings as a single input and utilize self-attention blocks, without a distinct process for handling the two different types of vectors.
In contrast, our approach treats the magnitude and syndrome as \emph{multimodal data}, recognizing their distinct informational characteristics.
The \emph{real-valued} magnitude vector contains the reliabilities of all bit positions, while the \emph{binary} syndrome vector conveys the information of erroneous bit positions. 
This deliberate separation necessitates the development of a novel architecture, specifically designed to effectively update these separated magnitude and syndrome embeddings, thereby significantly improving decoding performance.



In this paper, we introduce a novel Cross-attention Message-Passing Transformer~(CrossMPT) for ECC decoding.
CrossMPT processes the magnitude and syndrome separately to effectively utilize their distinct informational properties.
It employs two \emph{cross-attention blocks} to iteratively update the magnitude and syndrome embeddings. 
Initially, the magnitude embedding is encoded into the \emph{query}, while the syndrome embedding is encoded into \emph{key} and \emph{value}. 
The first cross-attention block utilizes this configuration in its attention mechanism to update the magnitude embedding component. 
This procedure is reciprocated for the syndrome embedding, which is encoded into the query, while the magnitude embedding is encoded into the key and value.
This configuration enables the second cross-attention block to update the syndrome vector component. 
These two masked cross-attention blocks iteratively collaborate to refine the magnitude and syndrome embeddings as in the message-passing algorithm~\citep{b_BP}.

To facilitate training, CrossMPT employs a mask matrix for each cross-attention block.
The first cross-attention block uses the transpose of the parity check matrix (PCM) $H^T$ as its mask matrix with the magnitude embedding as the query.
In the second cross-attention block, the PCM $H^{\top}$ itself is applied as the mask matrix, with the syndrome embedding acting as the query.
This strategy leverages the PCM's inherent representation of the `magnitude-syndrome' relationship, effectively aligning with the architecture's objectives.
Moreover, the combined size of the two attention maps of CrossMPT is at most half that of the attention map of the conventional transformer-decoder, leading to significantly reduced memory usage.
This reduction enables efficient learning and decoding of longer codes, which previous approaches~(concatenating magnitude and syndrome embeddings) are unable to achieve due to high memory usage and computational complexity.
To our knowledge, CrossMPT is the first architecture to integrate an iterative message-passing framework with a cross-attention-based transformer architecture.


Experimental results show that CrossMPT consistently outperforms the original ECCT across various code classes.
Leveraging its shared operational principles with the message-passing algorithm, CrossMPT demonstrates particularly improved decoding performance, especially in low-density parity-check~(LDPC) codes.
Notably, we also demonstrate that CrossMPT closely approaches the maximum likelihood decoding performance on short codes.
In addition to its enhanced decoding performance, CrossMPT significantly reduces the computational complexity (e.g., floating point operations~(FLOPs), training time, and inference time) of the decoder layer compared to the original ECCT.
Given that the decoder layer constitutes a substantial portion of the total computational cost, this reduction leads to a significant decrease in overall computational complexity.

\section{Related Works}\label{sec_ECCT_related_works}

In the field of neural network-based ECC decoders, there are two primary categories: the model-based decoder and the model-free decoder.
First, model-based decoders are constructed based on the conventional decoding methods~(e.g., BP decoder and MS decoder).
They map the iterative decoding process of the conventional decoding methods into neural networks and train the network weights accordingly.
To improve performance over the standard BP decoder, the recurrent neural network was employed for the decoding of BCH codes~\citep{b_Nachmani2018}.
Several recent studies showed that neural network-based BP and MS decoders outperform the conventional decoding algorithms over various code types~\citep{b_Dai2021, b_kwak2024, b_Lugosch2017, b_Nachmani2019, b_Nachmani2021, b_Kwak2022, b_Buchberger2021}.
However, model-based neural decoders may encounter performance limitations due to their restrictive model architectures, which are closely tied to underlying decoding methods.

Unlike model-based decoders, model-free neural decoders employ arbitrary neural network architectures to learn the decoding without relying on specific decoding algorithms.
Model-based neural decoders are constrained by the inherent limitations of the underlying decoding method (e.g., BP).
Consequently, they are unlikely to significantly exceed the performance limitations of traditional decoding algorithms.
On the other hand, model-free neural decoders can leverage state-of-the-art neural network architectures without such constraints.
Previous approaches~\citep{b_Gruber2017, b_Cammerer2017, b_Kim2018} implemented fully-connected network to decode codes but faced challenges in training due to overfitting.
Subsequently, the introduction of a preprocessing step utilizing the magnitude and syndrome vectors of the received codeword to learn multiplicative noise has been pivotal in enabling model-free decoders to address the overfitting issue~\citep{b_Bennantan2018}.
Then, ECCT~\citep{b_ECCT} first employed the transformer architecture using the same preprocessing step and demonstrated that the transformer-based decoder outperforms existing decoders including model-based neural decoders.
Building on the ECCT framework, denoising diffusion error correction codes~\citep{b_DDECC} interpreted the decoding process as a diffusion process and incorporated a diffusion model to train the original ECCT.
Recently, double-masked ECCT~\citep{b_Park2023} utilized two different PCMs for the same linear code to capture the diverse multilateral relationships of the magnitude and syndrome bits and improve the decoding performance.
Notably, transformer-based decoders outperform model-based neural decoders and serve as universal decoders capable of decoding arbitrary code classes with a unified architecture.

\section{Background}

\subsection{Error Correcting Codes}

Let $C$ be a linear block code, which is defined by a generator matrix $G$ of size $k\times n$ and a parity check matrix $H$ of size $(n-k)\times n$.
They satisfy $GH^{\top}=0$ over $\{0,1\}$ with modulo $2$ addition.
A codeword $x\in C \subset \{0,1\}^n$ is encoded by multiplying message $m$ with the generator matrix $G$~(i.e., $x=mG$). 
Let $x_s$ be the binary phase shift keying~(BPSK) modulated signal of $x$ and let $y$ be the output of a noisy channel for input $x_s$. We assume the additive white Gaussian noise~(AWGN) channel and the channel output can be represented by $y=x_s+z$, where \mbox{$z\sim N(0,\sigma^2)$}.
The objective of the decoder~($f:\mathbb{R}^n \rightarrow \mathbb{R}^n$) is to recover the transmitted codeword $x$ by correcting errors.
When $y$ is received, the decoder first determines whether the received signal is corrupted or not by checking the syndrome $s(y)=Hy_b$, where $y_b=\text{bin}(\text{sign}(y))$ is the demodulated signal of $y$. Here, $\text{sign}(a)$ represents $+1$ if $a\ge 0$ and $-1$ otherwise and $\text{bin}(-1)=1$, $\text{bin}(+1)=0$.
If $s(y)$ is a non-zero vector, it is detected that $y$ is corrupted during the transmission, and the decoder initiates the error correction process.

\begin{figure*}[!t]
\begin{center}
\subfigure[ECCT~\citep{b_ECCT}\label{fig_mask_a}]{\includegraphics[width=.48\textwidth]{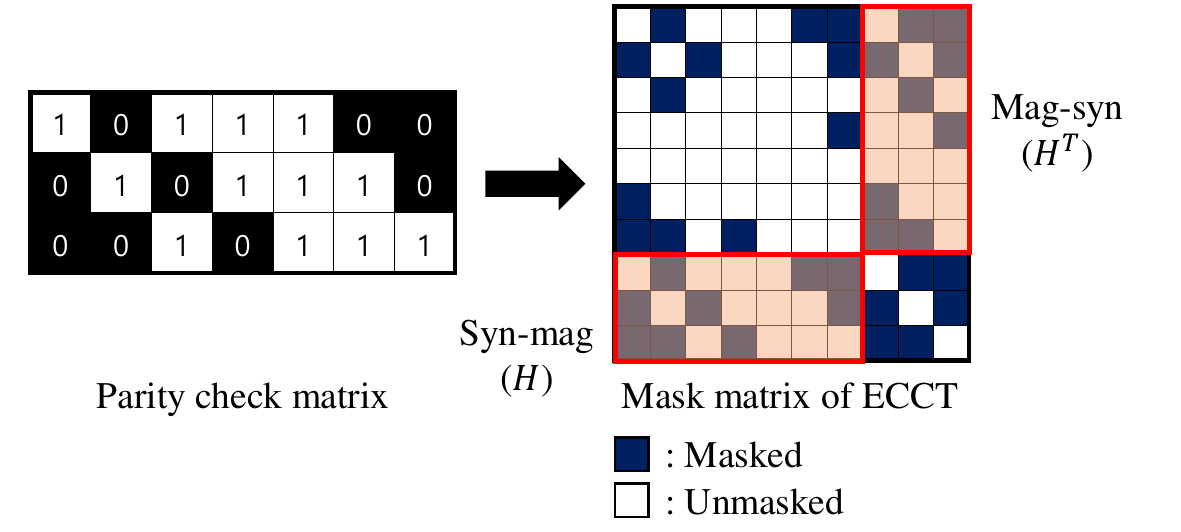}}
\subfigure[CrossMPT\label{fig_mask_b}]{\includegraphics[width=.48\textwidth]{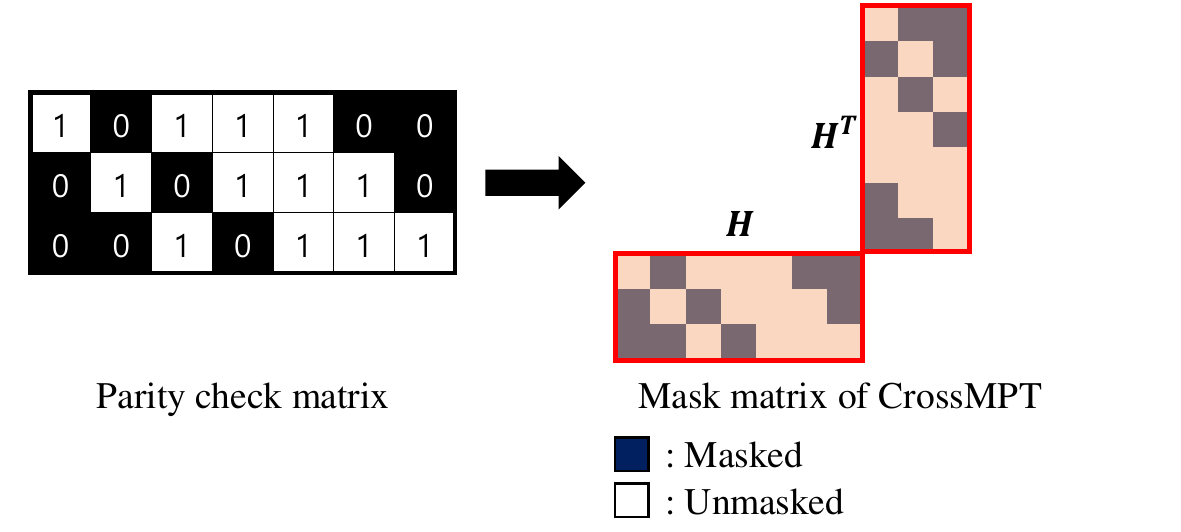}}
\caption{The PCM and the mask matrices of ECCT and CrossMPT\label{fig_mask}}
\end{center}
\end{figure*}

\begin{figure}[!t]
\begin{center}
\centerline{\includegraphics[width=.9\columnwidth]{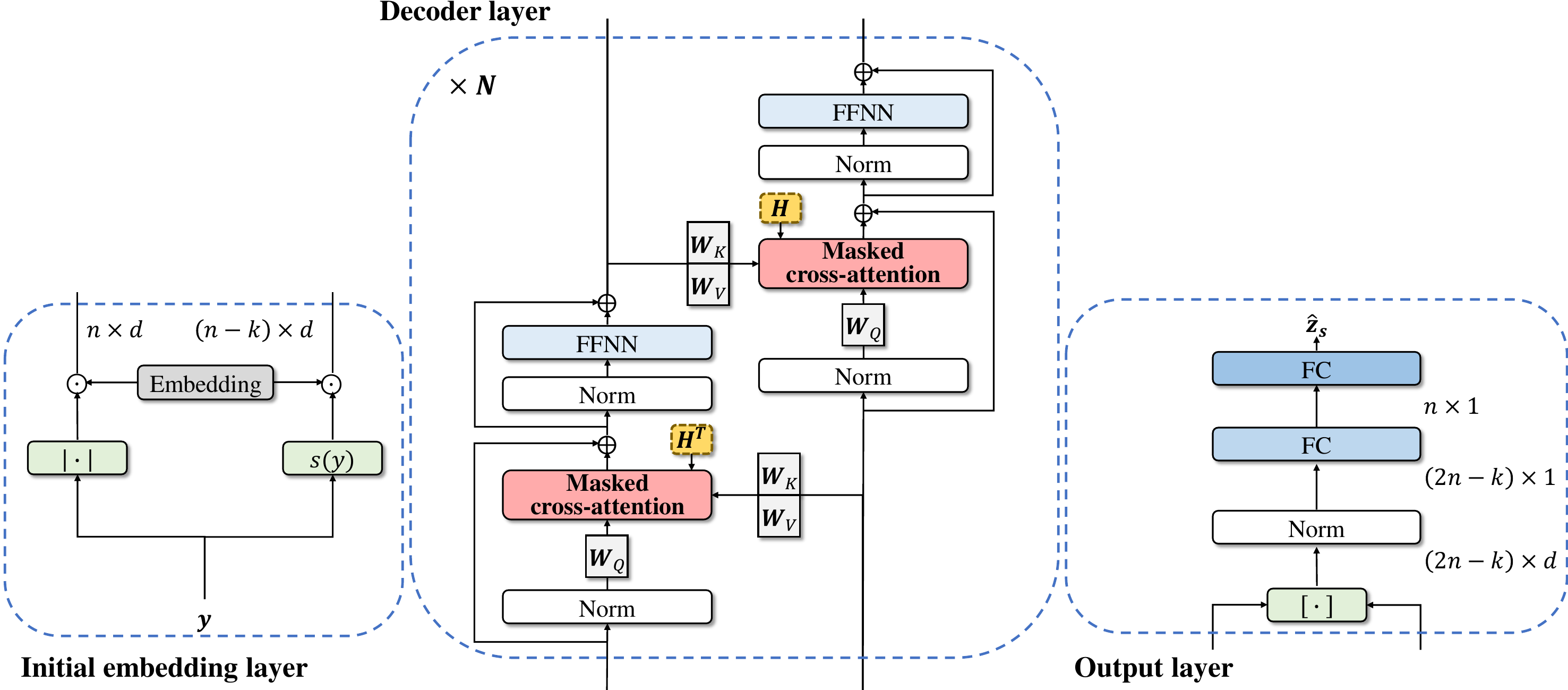}}
\caption{Architecture of CrossMPT.\label{fig_archi}}
\end{center}
\end{figure}

\subsection{Error Correction Code Transformer}
ECCT is the first approach to present a model-free decoder with the transformer architecture.
ECCT outperforms other neural BP-based decoders by employing the masked self-attention mechanism, whose mask matrix is determined by the code's PCM~\citep{b_ECCT}.
The first issue that needs to be addressed when training transformer-based decoders is the overfitting.
In \citep{b_Bennantan2018}, the overfitting issue in model-free neural decoders is described as a poor generalization to untrained new codewords due to the exponentially large number of codewords.
However, it has been resolved by a preprocessing technique that facilitates a syndrome-based decoding~\citep{b_Bennantan2018}. 
It has been theoretically proven that, with this preprocessing step, the decoder’s performance is invariant to the specific codewords in the training set~\citep{b_Bennantan2018}.

As in~\citep{b_Bennantan2018}, the preprocessing step of ECCT utilizes the magnitude and syndrome vectors to train multiplicative noise $\tilde{z}_s$, which is defined by
\begin{equation}
\label{equ_multi_noise}
    y=x_s+z = x_s\tilde{z}_s.
\end{equation}

ECCT aims to estimate the multiplicative noise in (\ref{equ_multi_noise}), i.e.,  \mbox{$f(y)=\hat{z}_s$}.
Then, the estimation of $x$ is
\mbox{$\hat{x} = \text{bin}(\text{sign}(yf(y)))$}.
If the multiplicative noise is correctly estimated such that \mbox{$\text{sign}(\tilde{z}_s)=\text{sign}(\hat{z}_s)$}, then $\hat{x}$ can be computed as:
\begin{equation*}
    \hat{x} = \text{bin}(\text{sign}(yf(y)))= \text{bin}(\text{sign}(x_s\tilde{z}_s\hat{z}_s))=\text{bin}(\text{sign}(x_s))=x.
\end{equation*}

ECCT employs a masked self-attention module to train the transformer architecture, where the input embedding is the concatenation of the embedded magnitude and syndrome vector of $y$.
As shown in Figure~\ref{fig_mask}, the mask matrice of ECCT should clearly distinguish between necessary (unmasked) and unnecessary (masked) pairwise relationships among magnitude-magnitude, magnitude-syndrome, and syndrome-syndrome bit relations.
In ECCT, the syndrome-syndrome part was only unmasked for self-relations, while the magnitude-syndrome part was unmasked based on the connections defined by the parity-check matrix (PCM).
The magnitude-magnitude part, however, was unmasked for bit pairs connected at depth 2 (see Algorithm 1 in \cite{b_ECCT}).
While the masking of magnitude-syndrome relations is intuitive, as it directly uses PCM, determining the relationships among magnitude themselves is not directly derivable from the PCM.
Therefore, the algorithm for masking magnitude-magnitude part is neither straightforward nor unique.
In Figure~\ref{fig_mask}, the white areas indicate unmasked positions (require the attention calculation) whereas the blue areas represent masked positions (omit the attention calculation).
As the proportion of blue increases, the attention matrix becomes sparser and more cost-efficient.


\section{Cross-attention Message-Passing Transformer} \label{sec:methods}

In this section, we present the operational mechanism and architecture of CrossMPT.
CrossMPT handles the magnitude and syndrome embeddings separately, applying a cross-attention mechanism to effectively capture their distinct information.
It shares its core principles with message passing decoding algorithm for decoding linear codes, where the magnitude and syndrome embeddings of the received codewords are iteratively updated.
The overall architecture is illustrated in Figure~\ref{fig_archi}.

\subsection{Cross-attention Message-passing Transformer}


One cross-attention block updates the magnitude embedding by using it as a query and updates them with key and value, generated from the syndrome embedding.
Given this configuration, the attention map has the size $n\times (n-k)$, effectively representing the `magnitude-syndrome' relation.
To reflect this relationship, we employ the transpose of the PCM $H^{\top}$ as the mask matrix.
This is because the $n$ rows of $H^{\top}$ correspond to the $n$ bit positions, and its $n-k$ columns are associated with the parity check equations, directly linking to $|y|$ and $s(y)$, respectively.
The other cross-attention block similarly use $s(y)$ is used as the query, while $|y|$ serves as both key and value.
For this operation, we utilize the PCM $H$ as the mask matrix.

This configuration of separately handling two distinct informational properties resembles the message-passing decoding algorithm for decoding linear codes.
Message-passing algorithms such as the sum-product algorithm~\citep{b_BP} are widely used for decoding ECCs due to their outstanding decoding performance with low complexity.
The message-passing algorithm operates by exchanging messages between variable nodes~(VNs) and check nodes~(CNs) over a Tanner~(bipartite) graph~\citep{b_BP}.
In the Tanner graph, VNs convey information about the reliability of the received codeword, while CNs indicate the parity check equations.
The edges between VNs and CNs represent the connections~(relationships) between them.
The message-passing decoder operates by exchanging messages between VNs and CNs via edges.
The output messages of VNs and CNs are updated in an iterative manner.

Similar to the principles of message-passing algorithms, CrossMPT updates $|y|$ and $s(y)$ by allowing them to exchange messages with each other.
Initially, we update $|y|$ by the masked cross-attention block, with $|y|$ as the query and $s(y)$ as both the key and value.
The syndrome embedding is updated in the subsequent masked cross-attention block, utilizing the previously updated magnitude embedding.
In this block, the syndrome embedding is used as the query, while the updated magnitude embedding serves as the key and value.
The resulting output from this cross-attention block is the updated syndrome.
CrossMPT iteratively updates both the magnitude and syndrome embeddings to identify the multiplicative noise accurately.

\begin{figure}[!t]
\begin{center}
\centerline{\includegraphics[width=.75\columnwidth]{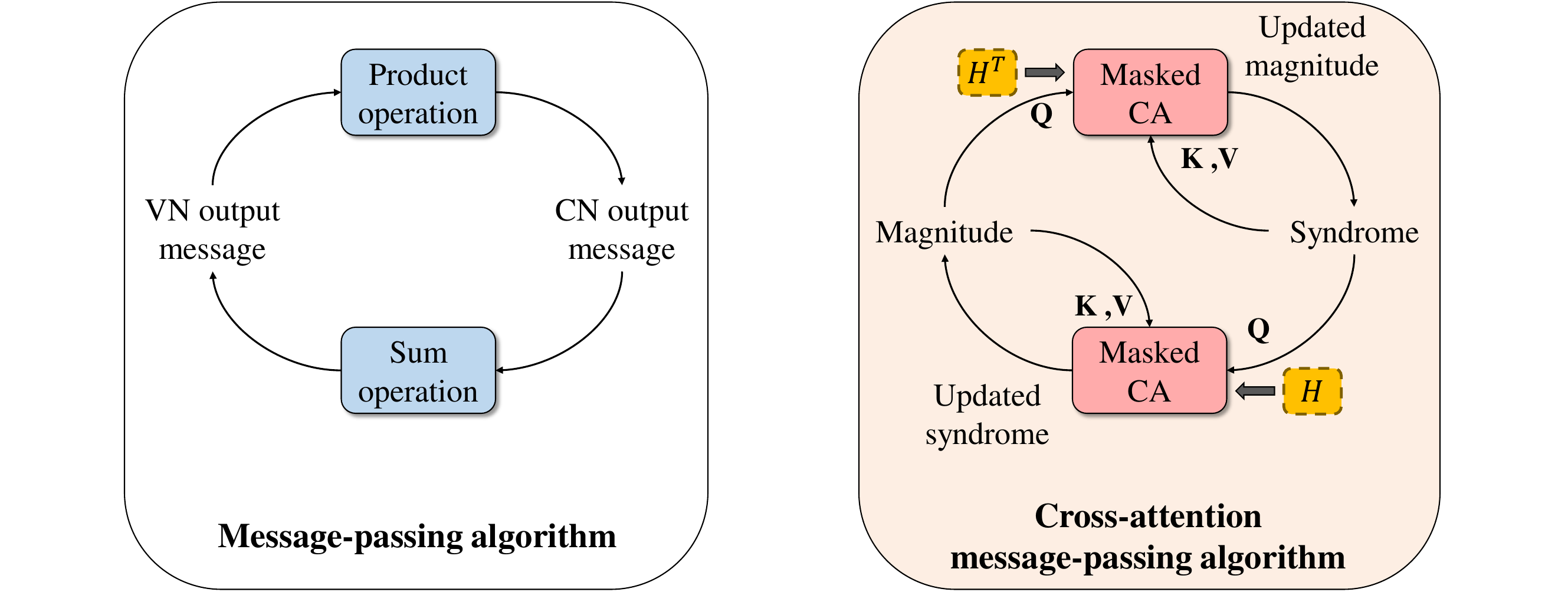}}
\caption{Conceptual comparison of the sum-product message-passing algorithm and the proposed cross-attention~(CA) message-passing algorithm.\label{fig_message_passing}}
\end{center}
\end{figure}

As a representative of the message-passing algorithm, Figure~\ref{fig_message_passing} depicts the sum-product algorithm and the cross-attention message-passing algorithm.
In the sum-product algorithm, the VN output and CN output messages are iteratively updated using the sum and product operations.
Similar to the sum-product algorithm, the magnitude and syndrome embeddings are iteratively updated using the masked cross-attention~(Masked CA in the figure) blocks in CrossMPT.
Note that $H$ and $H^{\top}$ are utilized for the mask matrices for these cross-attention blocks, and ${Q}$, ${K}$, and ${V}$ represent the query, key, and value of the cross-attention mechanism.

\subsection{Model Architecture}

In the initial embedding layer, we generate $|y|=(|y_1|,\ldots,|y_n|)$ and $s(y)=(s(y)_1,\ldots ,s(y)_{n-k})$ from the received codeword, and project each element $y_i$ and $s(y)_i$ into $d$ dimension embedding row vectors $M_i$ and $S_i$, respectively, as follows:
\begin{align*}
    &M_i=|y_i|W_i,\qquad~~\text{ for } i=1,\ldots,n,\\
    &S_i=s(y)_iW_{i+n},\quad \text{ for } i=1,\ldots,n-k,
\end{align*}
where $W_i \in \mathbb{R}^{1\times d}$ for $i=1,\ldots,2n-k$ denote the trainable positional encoding vector.

These two embedded vectors are processed as separate input vectors in the following $N$ decoding layers.
Each decoding layer contains two cross-attention blocks, each consisting of a cross-attention module, a feed-forward neural network (FFNN), and a normalization layer.

In the first cross-attention module, the attention module updates the `magnitude' embedding by using the syndrome.
The query vector $Q_1$, key vector $K_1$, and value vector $V_1$ are assigned as follows:
\begin{equation*}
Q_1=MW_Q, K_1=SW_K, V_1=SW_V,
\end{equation*}
where $M=[ M_1; \cdots; M_n]\in\mathbb{R}^{n\times d}$ and $S=[ S_1; \cdots ; S_{n-k}]\in\mathbb{R}^{(n-k)\times d}$ denote the magnitude embedding and the syndrome embedding, respectively, and $W_Q,W_K,W_V$ denote the weight matrices of query, key, and value, respectively.
This architecture is termed the `cross-attention' message-passing transformer since the query corresponds to the magnitude embedding, while the key and value correspond to the syndrome embedding. Then, we employ the following scaled dot-product attention:
\begin{equation*}
{\rm Attention}(Q_1,K_1,V_1)={\rm softmax}\left(\dfrac{Q_1K_1^{\top}+g(H^{\top})}{\sqrt{d}}\right)V_1,
\end{equation*}
where $g(H^{\top})$ is the mask matrix, and the function $g$ is defined as

\begin{equation}
    g(A)_{i,j}=\begin{cases} 0 \qquad~{\rm if }~A_{i,j}=1,\\ -\infty\quad{\rm if }~A_{i,j}=0. \end{cases}
\end{equation}

This configuration results in an attention map of size $n\times(n-k)$, representing the ‘magnitude-syndrome’ relationship.
Therefore, we use the transpose of the PCM $H^{\top}$ as a mask matrix since the $n$ rows of $H^{\top}$ correspond to the $n$ bit positions and the $n-k$ columns of $H^{\top}$ to the parity check equations, which are closely related to $|y|$ and $s(y)$, respectively.
Finally, the output vector embodies a newly updated magnitude embedding $M'$ by using the syndrome.

In the second cross-attention module, we update the `syndrome' embedding with the updated $M’$ corresponding to the magnitude. In other words, the input of the query becomes syndrome and the input of key and value becomes $M’$. We use the \textit{shared weight vectors} $W_Q,W_K,W_V$ as in the first cross-attention module, and query vector $Q_2$, key vector $K_2$, and value vector $V_2$ are defined as follows:
\begin{equation*}
    Q_2=SW_Q, K_2=M'W_K, V_2=M'W_V.
\end{equation*}

Here, the syndrome and magnitude correspond the row and column of the attention map, respectively.
Thus, we employ the mask matrix $g(H)$, whose masking positions are zeros in $H$. Then we apply the scaled dot-product attention and the resulting output vector conveys the updated syndrome.
This output vector is utilized to further refine the magnitude embedding and this process is iteratively repeated across the $N$ decoder layers.

Finally, these output vectors of the last decoder layer are concatenated and pass through a normalization layer and two fully connected~(FC) layers.
The first FC layer reduces the $(2n-k) \times d$ dimension embedding to a one-dimensional $2n-k$ vector, and the second FC layer further reduces the dimension from $2n-k$ into $n$.
The final output provides an estimation of $\tilde{z}_s$.
Since two cross-attention blocks of CrossMPT share the same weight matrices $W_Q, W_K, W_V$ and all other layers, CrossMPT has the same number of parameters as the original ECCT.

\subsection{Training}

The objective of the proposed decoder is to learn the multiplicative noise $\tilde{z}_s$ in (\ref{equ_multi_noise}) and reconstruct the original transmitted signal $x$.
We can obtain the multiplicative noise by $\tilde{z}_s = \tilde{z}_s x^2_s = yx_s$.
Then, the target multiplicative noise for binary cross-entropy loss function is defined by $\tilde{z} =\text{bin}(\text{sign}(yx_s))$.
Finally, the cross-entropy loss function for a received codeword $y$ is defined by
\begin{equation*}
    \mathcal{L} = -\sum^{n}_{i=1}\left\{\tilde{z}_i \log (\sigma(f(y))) + (1-\tilde{z}_i) \log(1-\sigma(f(y))) \right\}.
\end{equation*}

To ensure a fair comparison between CrossMPT and ECCT, we adopt the same training setup used in the previous work~\citep{b_ECCT}.
We use the Adam optimizer~\citep{b_adam} and conduct 1000 epochs.
Each epoch consists of 1000 minibatches, where each minibatch is composed of 128 samples.
All simulations were conducted using NVIDIA GeForce RTX 3090 GPU and AMD Ryzen 9 5950X 16-Core Processor CPU.
The training sample $y$ is generated by $y=x_s+z$, where  $x_s$ is the all-zero codeword and the AWGN channel noise $z$ is from an SNR~($E_b/N_0$) range of $3$~dB to $7$~dB.
The learning rate is initially set to $10^{-4}$ and gradually reduced to $5\times10^{-7}$ following a cosine decay scheduler.

\begin{table*}[!t]
\caption{Comparison of decoding performance at three different SNR values (4 dB, 5 dB, 6 dB) for BP decoder, Hyper BP decoder~\citep{b_Nachmani2019}, AR BP decoder~\citep{b_Nachmani2021}, ECCT~\citep{b_ECCT}, and the proposed CrossMPT. The results are measured by the negative natural logarithm of BER. The best results are highlighted in \textbf{bold}. Higher is better.}
\label{tab_results}
\begin{center}
\begin{small}
\resizebox{\textwidth}{!}{
\begin{tabular}{ccccccccccccccccc}
\toprule
\multicolumn{2}{c}{Architecture} & \multicolumn{9}{c}{BP-based decoders} & \multicolumn{6}{c}{Model-free decoders}\\
\cmidrule(r){1-2}\cmidrule(r){3-11}\cmidrule(r){12-17}
\multirow{2}{*}{Codes} & \multirow{2}{*}{Parameter} & \multicolumn{3}{c}{BP} & \multicolumn{3}{c}{Hyp BP} & \multicolumn{3}{c}{AR BP} & \multicolumn{3}{c}{ECCT} & \multicolumn{3}{c}{CrossMPT} \\
\cmidrule(r){3-5}\cmidrule(r){6-8}\cmidrule(r){9-11}\cmidrule(r){12-14}\cmidrule(r){15-17}
&                         & 4     & 5     & 6      & 4      & 5       & 6       & 4      & 5       & 6      & 4      & 5      & 6      & 4       & 5        & 6       \\
\midrule
\multirow{3}{*}{BCH}   & (31,16)                 & 4.63  & 5.88  & 7.60   & 5.05   & 6.64    & 8.80    & 5.48   & 7.37    & 9.60   & 6.39	&8.29	&10.66	&\textbf{6.98}	&\textbf{9.25}	&\textbf{12.48}   \\
& (63,36)   & 4.03 &	5.42&	7.26&	4.29&	5.91	&8.01&	4.57&	6.39&	8.92  &	4.86	&6.65	&9.10	&\textbf{5.03}	&\textbf{6.91}	&\textbf{9.37}\\
& (63,45)   & 4.36  & 5.55  & 7.26   & 4.64   & 6.27    & 8.51    & 4.97   & 6.90    & 9.41   & 5.60	&7.79	&10.93	&\textbf{5.90}	&\textbf{8.20}	&\textbf{11.62}\\
& (63,51)   & 4.5	&5.82	&7.42	&4.8	&6.44	&8.58	&5.17	&7.16	&9.53	&5.66	&7.89	&11.01	&\textbf{5.78}	&\textbf{8.08}	&\textbf{11.41}\\
\midrule
\multirow{5}{*}{Polar} & (64,32)                 & 4.26  & 5.38  & 6.50   & 4.59   & 6.10    & 7.69    & 5.57   & 7.43    & 9.82   & 6.99	&9.44	&12.32  & \textbf{7.50}    & \textbf{9.97}     & \textbf{13.31}   \\
& (64,48)                 & 4.74  & 5.94  & 7.42   & 4.92   & 6.44    & 8.39    & 5.41   & 7.19    & 9.30   & 6.36	&8.46	&11.09  & \textbf{6.51}    & \textbf{8.70}     & \textbf{11.31}   \\
& (128,64)    & 4.1	&5.11	&6.15	&4.52	&6.12	&8.25	&4.84	&6.78	&9.3	&5.92	&8.64	&12.18	&\textbf{7.52}	&\textbf{11.21}	&\textbf{14.76} \\
& (128,86)     & 4.49	&5.65	&6.97	&4.95	&6.84	&9.28	&5.39	&7.37	&10.13	&6.31	&9.01	&12.45	&\textbf{7.51}	&\textbf{10.83}	&\textbf{15.24} \\
& (128,96)     & 4.61	&5.79	&7.08	&4.94	&6.76	&9.09	&5.27	&7.44	&10.2	&6.31	&9.12	&12.47	&\textbf{7.15}	&\textbf{10.15}	&\textbf{13.13} \\
\midrule
\multirow{4}{*}{LDPC}  & (49,24)                 & 6.23  & 8.19  & 11.72  & 6.23   & 8.54    & 11.95   & 6.58   & 9.39    & 12.39  & 6.13	&8.71	&12.10  & \textbf{6.68}    & \textbf{9.52}     & \textbf{13.19}   \\
& (121,60)                & 4.82  & 7.21  & 10.87  & 5.22   & 8.29    & 13.00   & 5.22   & 8.31    & 13.07  & 5.17   & 8.31   & 13.30  & \textbf{5.74}    & \textbf{9.26}     & \textbf{14.78}   \\
& (121,70)                & 5.88  & 8.76  & 13.04  & 6.39   & 9.81    & 14.04   & 6.45   & 10.01   & 14.77  & 6.40   & 10.21  & 16.11  & \textbf{7.06}    & \textbf{11.39}    & \textbf{17.52}   \\
& (121,80)                & 6.66  & 9.82  & 13.98  & 6.95   & 10.68   & 15.80   & 7.22   & 11.03   & 15.90  & 7.41   & 11.51  & 16.44  & \textbf{7.99}    & \textbf{12.75}    & \textbf{18.15}   \\
\midrule
MacKay                 & (96,48)                 & 6.84  & 9.40  & 12.57  & 7.19   & 10.02   & 13.16   & 7.43   & 10.65   & 14.65  & 7.38	&10.72	&14.83  & \textbf{7.97}    & \textbf{11.77}    & \textbf{15.52}   \\
\midrule
CCSDS                   & (128,64)                & 6.55  & 9.65  & 13.78  & 6.99   & 10.57   & 15.27   & 7.25   & 10.99   & 16.36  & 6.88   & 10.90  & 15.90  & \textbf{7.68}    & \textbf{11.88}    & \textbf{17.50}   \\
\midrule
Turbo                   & (132,40)                & N/A  & N/A  & N/A  & N/A   &  N/A   &  N/A   &  N/A   &  N/A   & N/A  & 4.74	&6.54	&9.06	&\textbf{5.55}	&\textbf{7.92}	&\textbf{10.94}
   \\
\bottomrule
\end{tabular}
}
\end{small}
\end{center}
\end{table*}

\begin{figure}[!t]
\begin{center}
\subfigure[$(31,16)$~BCH code]{\includegraphics[width=.32\textwidth]{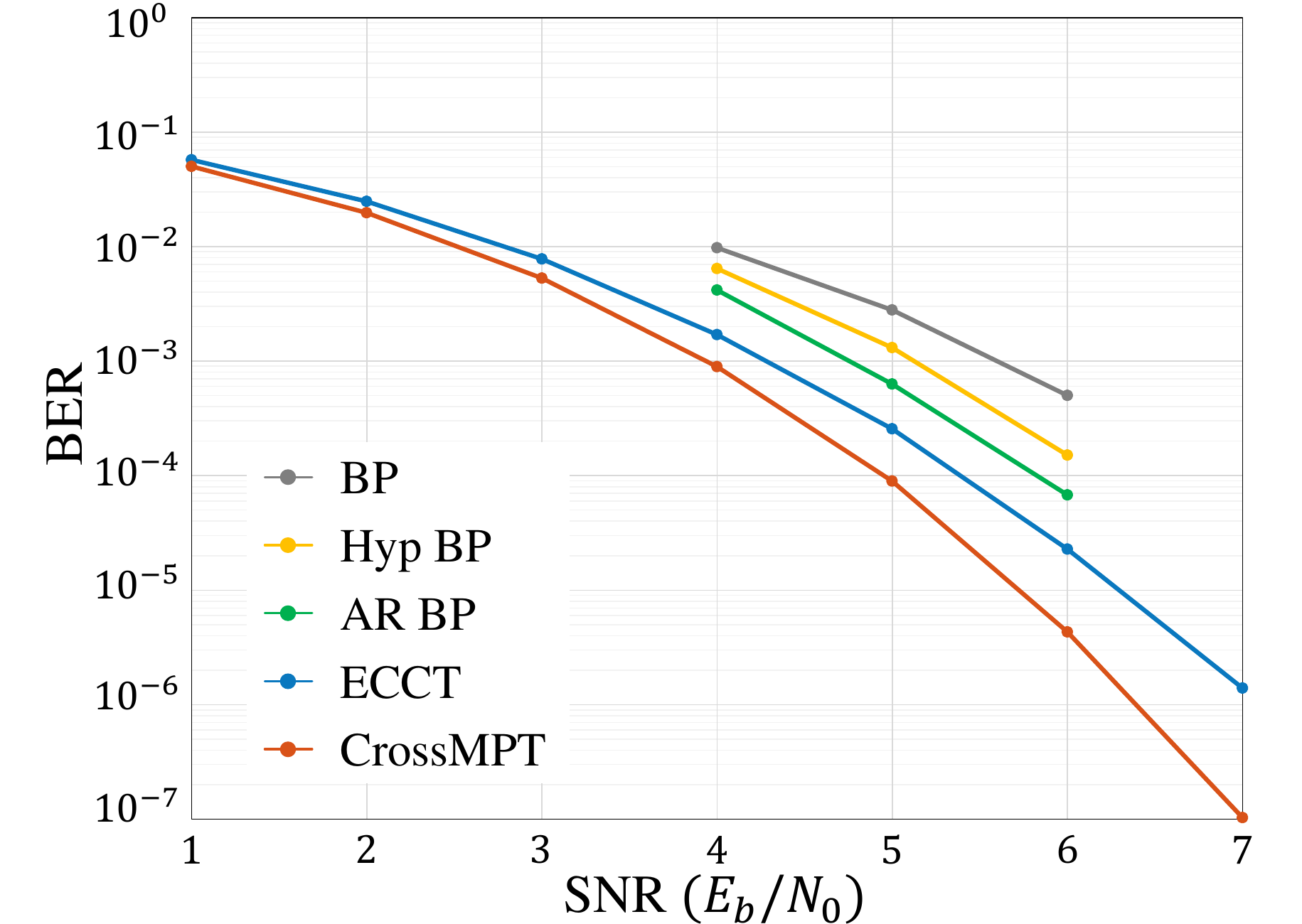}}
\subfigure[$(128,86)$~polar code]{\includegraphics[width=.32\textwidth]{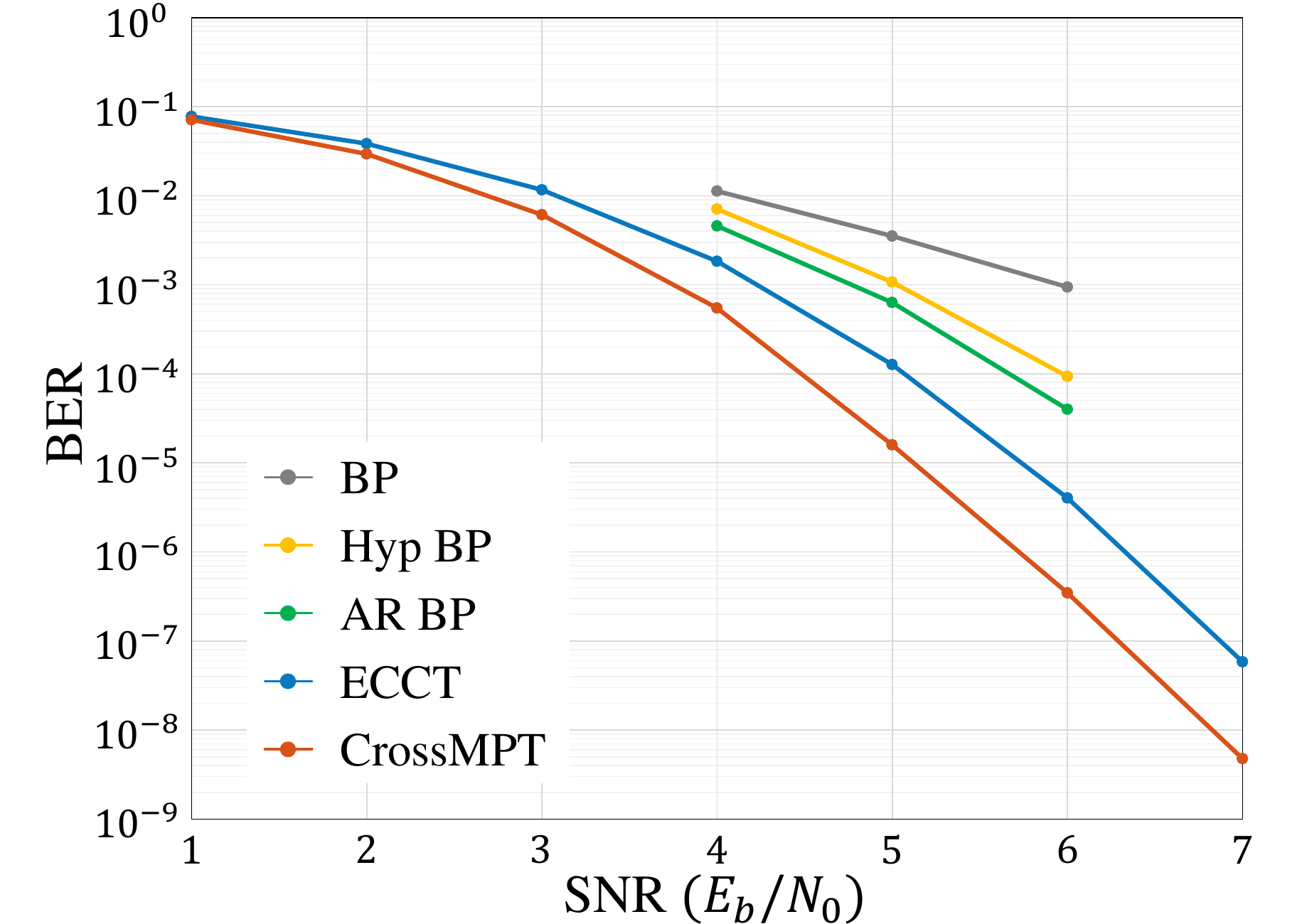}}
\subfigure[$(128,64)$~CCSDS]{
    \includegraphics[width=.32\textwidth]{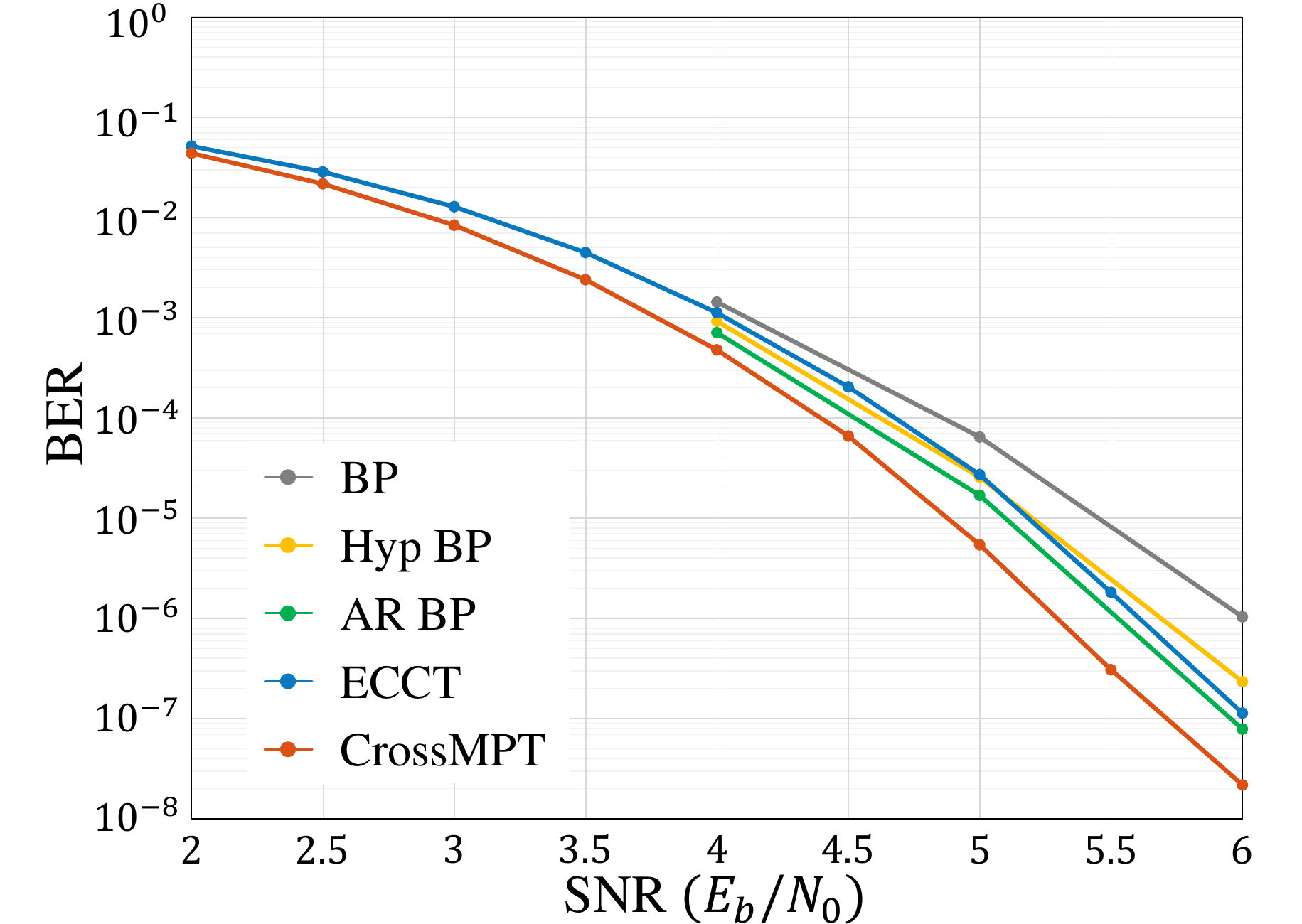}}
\caption{The BER performance of various decoders~(BP, Hyp BP, AR BP, ECCT) and CrossMPT.
\label{fig_graph}}
\end{center}
\end{figure}

\section{Experimental Results}

In this section, we compare the proposed CrossMPT with the original ECCT across various code classes.
Our experimental results do not include a comparison with the works of \citep{b_Choukroun2024,b_Choukroun2024_ICML}, as they have different objectives, such as generalizing the decoder to unseen codes \citep{b_Choukroun2024} or jointly training the encoder and decoder \citep{b_Choukroun2024_ICML}.
It is worth mentioning that our cross-attention architecture and the schemes of \citep{b_Choukroun2024,b_Choukroun2024_ICML} are orthogonal methods, and combining them could present a promising direction for future research.

To verify the efficacy of CrossMPT, we train it for BCH codes, polar codes, turbo codes, and LDPC codes~(including MacKay and  CCSDS codes) and evaluate the bit error rate~(BER) performance.
All PCMs are taken from~\citep{b_channelcode}.
The implementation of the original ECCT is taken from \citep{b_ECCT_code}.
For the testing, we collect at least 500 frame errors at each signal-to-noise ratio~(SNR) value with random codewords.
Table~\ref{tab_results} compares the decoding performance of CrossMPT with the BP decoder, BP-based neural decoders~\citep{b_Nachmani2019,b_Nachmani2021}, and ECCT~\citep{b_ECCT}.
The results of the BP-based decoders in Table~\ref{tab_results} are obtained for 50 iterations.
The results for both the proposed CrossMPT and ECCT, which are model-free decoders, are obtained with $N=6$ and $d=128$.
For all types of codes, CrossMPT outperforms the conventional ECCT and all the other BP-based neural decoders.
This improvement of CrossMPT is particularly notable in the case of LDPC codes.
To provide more visual information, we plot the BER graphs for several codes in Figure~\ref{fig_graph}.

An important aspect of our research is CrossMPT's capability to decode long codes~(Appendix~\ref{append_long}), which remain beyond the reach of ECCT due to its high memory requirements, resulting from large attention maps.
These results demonstrate the practical significance and architectural advantages of CrossMPT, proving its value in scenarios where ECCT encounters limitations.
Especially, it achieves superior decoding performance for LDPC codes, outperforming the BP decoder with the maximum iteration of 100~(provided in Appendix~\ref{append_BP}).
Also, we demonstrate that for short codes, CrossMPT closely approaches the optimal maximum likelihood~(ML) decoding performance~(provided in Appendix~\ref{append_ML}).
Additional experimental results for comparison with successive cancellation list polar decoder, denoising diffusion ECCT~(DDECCT), and the decoding performance for the Rayleigh channel are provided in Appendices~\ref{append_SCL}, \ref{append_DDECCT}, and \ref{append_Rayleigh}, respectively.

\begin{figure}[!t]
\centering
\includegraphics[width=.6\linewidth]{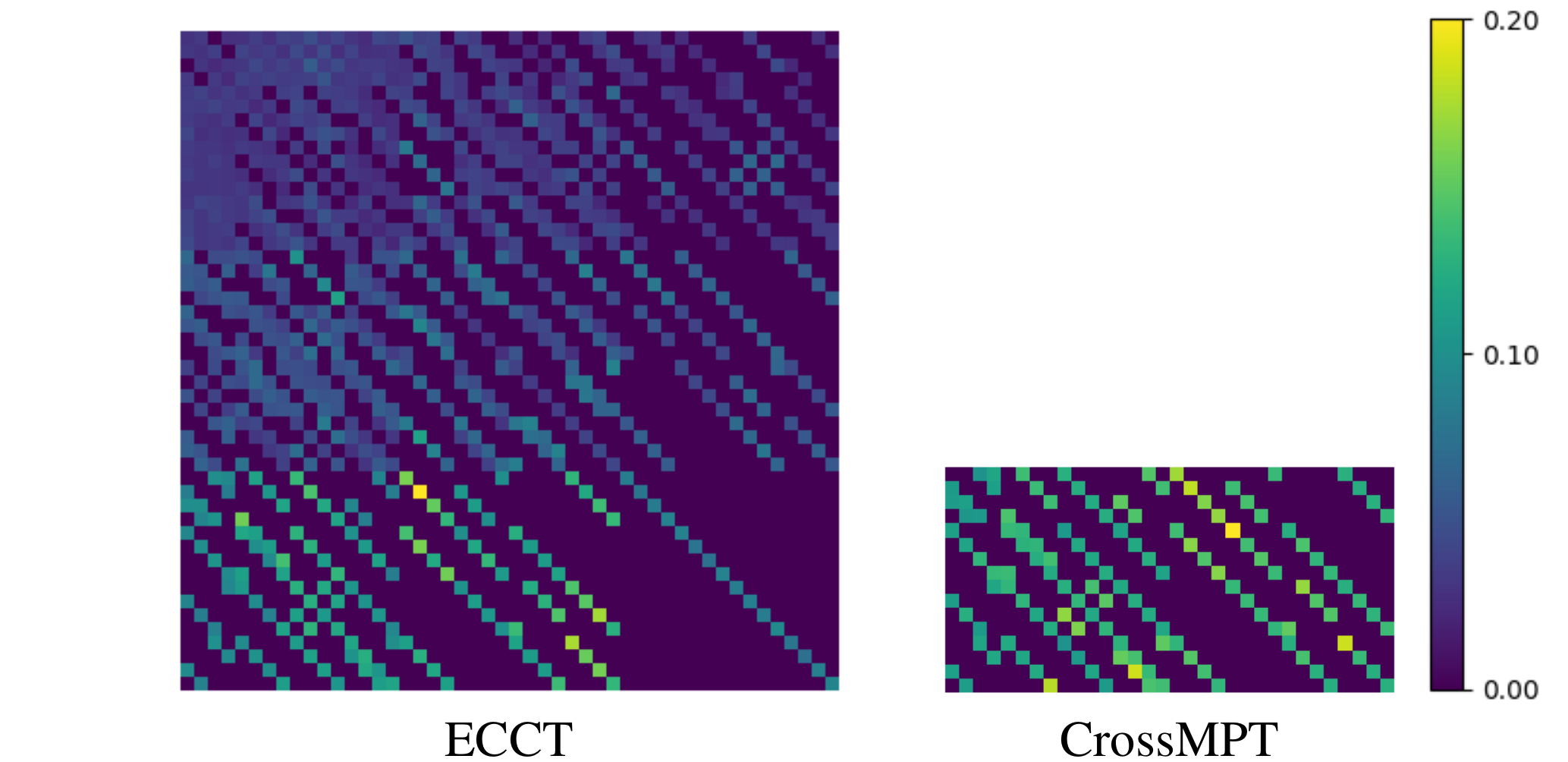}
\caption{The average attention scores of all $N=6$ layers for ECCT and CrossMPT.\label{fig_attn_avg}}
\end{figure}

\section{Ablation Studies and Analysis}

\subsection{Analysis of Attention Mechanisms in ECCT and CrossMPT}

We provide a comparative analysis of the attention scores in ECCT and CrossMPT.
Figure~\ref{fig_attn_avg} shows the average attention scores across $N=6$ layers for both ECCT and CrossMPT for $(32,16)$~LDPC code~\citep{b_Divsalar2010}.
As shown in Figure~\ref{fig_attn_avg}, the attention score map of ECCT reveals different importance among the relationships: magnitude-magnitude, syndrome-syndrome, and magnitude-syndrome.
One key observation is that the magnitude-magnitude and syndrome-syndrome relations exhibit relatively low attention scores compared to the magnitude-syndrome relation in Figure~\ref{fig_attn_avg}.
This suggests that the magnitude-syndrome relationship is more significant than the others.
An ablation study, in which we masked the magnitude-magnitude and syndrome-syndrome relationships, revealed no significant performance difference compared to when these relationships were not masked (see Appendix~\ref{append_ablation_masking}).
This ablation study shows that the conventional ECCT could be enhanced by focusing on the more critical relationships.
However, as shown in Figure~\ref{fig_attn_avg}, CrossMPT eliminates the two relations with low attention scores and focuses only on the magnitude-syndrome relation.
Therefore, we can claim that CrossMPT more efficiently targets the crucial aspect (i.e., magnitude-syndrome relation) compared to ECCT.

\begin{figure}[!t]
\begin{center}
\subfigure[Attention scores with a single bit error \label{fig_attn_score_a}]{\includegraphics[width=.6\textwidth]{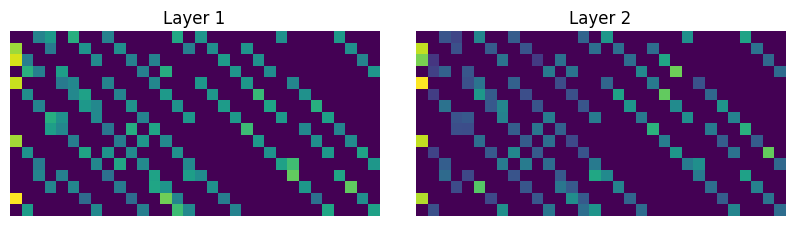}}
\hfill
\subfigure[Summation of attention scores with a single bit error \label{fig_attn_score_b}]{\includegraphics[width=.3\textwidth]{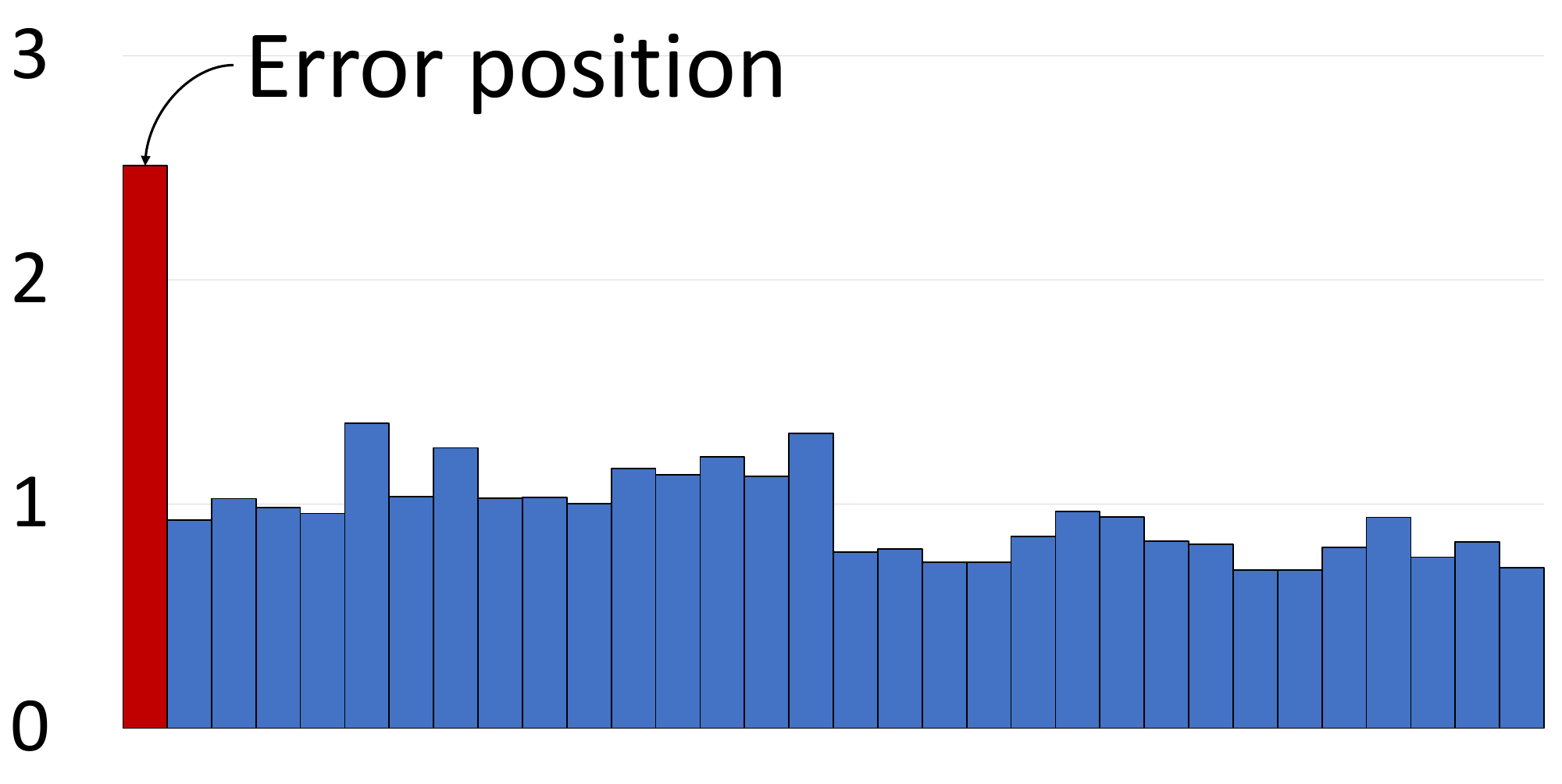}}
\subfigure[Attention scores without an error \label{fig_attn_score_c}]{\includegraphics[width=.6\textwidth]{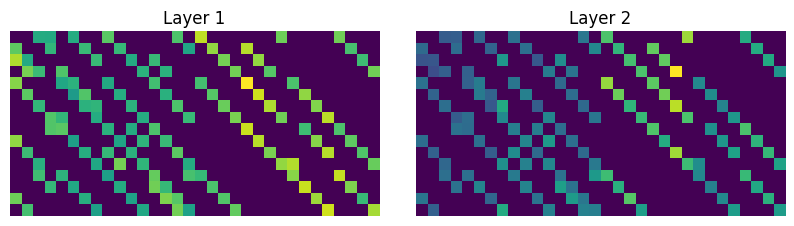}}
\hfill
\subfigure[Summation of attention scores without an error \label{fig_attn_score_d}]{\includegraphics[width=.3\textwidth]{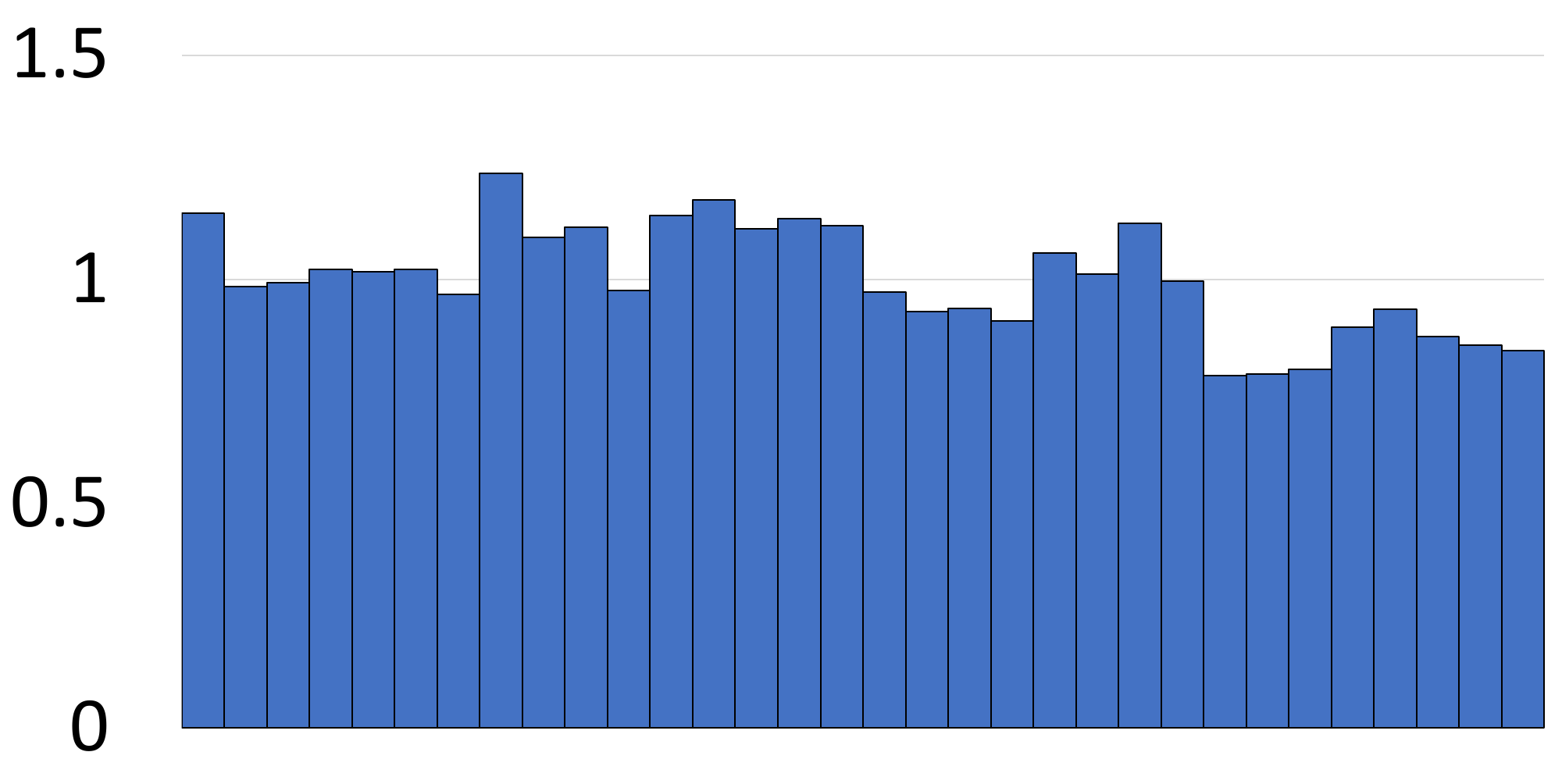}}
\caption{The attention scores (a), (c) with a single bit error in the first bit position and without an error.
The summation of the attention scores (b), (d) is carried out in the vertical direction.
\label{fig_attn_score}}
\end{center}
\end{figure}

\subsection{Visualization of Cross-attention Map}
To further examine how CrossMPT operates, we intentionally corrupt a pre-determined bit of the $(32,16)$~LDPC code and analyze the resulting attention maps.
Figure~\ref{fig_attn_score} shows the attention scores for the first two layers and the summation of their attention scores when the \textit{first bit} is corrupted.
The summation is carried out vertically to demonstrate the attention score for each bit.
As shown in Figure~\ref{fig_attn_score}(b), the attention score of the first bit~(or first column) is relatively higher than the others.
However, once the error is corrected, CrossMPT no longer assigns high scores to that position~(see Figure~\ref{fig_append_attn_avg} in Appendix~\ref{append_attn_avg}).
Figures~\ref{fig_attn_score_c} and \ref{fig_attn_score_d} depict the attention scores and the summation of attention scores without an error.
Compared to the previous case, the attention scores are more uniformly distributed across all positions.

\subsection{Number of Parameters}
Two cross-attention blocks of CrossMPT share the same parameters for all decoder layers.
They use the same weight matrices $W_Q, W_K, W_V$ for two cross-attention modules since the performance remains nearly identical even when the parameters are trained separately.
Also, they share the parameters for the normalization layer and the FFNN layer.
Thus, CrossMPT has the same number of parameters as the original ECCT.

\subsection{Complexity Analysis}

\begin{figure}[!t]
\begin{center}
\centerline{\includegraphics[width=.8\columnwidth]{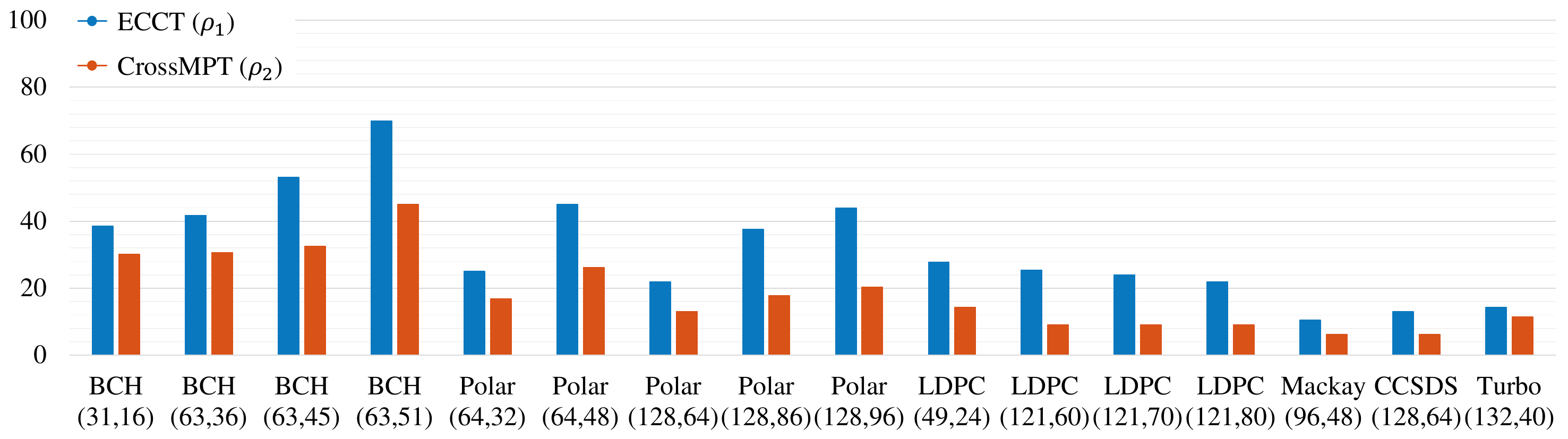}}
\caption{Comparison of the mask matrix density between ECCT and CrossMPT.\label{fig_sparsity}}
\end{center}
\end{figure}

Figure~\ref{fig_mask} illustrates the mask matrices of ECCT and CrossMPT.
In the original ECCT, Figure~\ref{fig_mask}(a) shows that a significant portion of the upper $n \times n$ submatrix is depicted in white, indicating that the most positions are unmasked.
This $n \times n$ submatrix represents depth-2 connections in the Tanner graph~\citep{b_ECCT}, which results in an increase in the number of unmasked positions, thereby leading to a higher computational required.
On the other hand, the lower $(n-k) \times n$ submatrix and the right $(n-k) \times n$ submatrix, which serve as the masking matrices for CrossMPT, are predominantly shown in blue, indicating that their attention matrices are sparser.
Figure~\ref{fig_sparsity} compares the mask matrix density of CrossMPT and ECCT.
For all codes, the mask matrix of CrossMPT is sparser than ECCT, which implies that CrossMPT can achieve lower computational complexity compared to the original ECCT. 

The complexity of the self-attention mechanism of ECCT, without considering the masking is, $\mathcal{O}(N(d^2(2n-k)+(2n-k)^2d))$.
When taking masking into account, the complexity can be reduced to $\mathcal{O}(N(d^2(2n-k)+hd))$~\citep{b_ECCT}, where $h = \rho_{1}(2n-k)^2$ denotes the fixed number of computations of the self-attention module and $\rho_{1}$ denotes the density of the mask matrix in ECCT.
Similarly, the complexity of the two cross-attention modules of CrossMPT, without considering the masking, is $\mathcal{O}(N(d^2(2n-k)+2n(n-k)d))$.
When masking is taken into account, the complexity can be reduced to $\mathcal{O}(N(d^2(2n-k)+(h_1 + h_2)d))$, where $h_1 = \rho_{2} n(n-k)$ denotes the number of computations of the first cross-attention module, $h_2 = \rho_{2} (n-k)n$ denotes the number of computations of the second cross-attention module, and $\rho_{2}$ denotes the density of the mask matrix in CrossMPT.
Furthermore, since $\rho_1 > \rho_2$ as shown in Figure~\ref{fig_sparsity}, we conclude that $h > h_1 + h_2$, which indicates that CrossMPT achieves a reduction in computational complexity compared the original ECCT.

Table~\ref{tab_time} compares the FLOPs, inference time, and training time between ECCT and CrossMPT.
The inference time refers to the duration required to decode a single codeword and the training time measures the duration to complete one epoch of training.
All results are obtained for $N=6$ and $d=128$, except for (255,223)~BCH code and (384,320) WRAN LDPC code, which are obtained for $N=6$ and $d=32$.
For all three metrics, CrossMPT outperforms ECCT.
Since the inference time and the training time are closely related to the FLOPs, a reduction in FLOPs directly leads to shorter inference and training times.
Notably, for long codes, CrossMPT achieves a significant reduction compared to ECCT.
The results in Tables~\ref{tab_results} and \ref{tab_time} demonstrate that the proposed CrossMPT not only improves the decoding performance but also significantly reduces FLOPs, inference time, and training time compared to the original ECCT.
Additional analysis on training convergence and throughput of CrossMPT is provided in Appendix~\ref{append_conv} and~\ref{append_throughput}, respectively.

\begin{table}[!t]
\caption{Comparison of FLOPs, inference time, and training time between ECCT and CrossMPT for various codes.
Inference time is measured for decoding a single codeword and training time is measured for a single epoch.\label{tab_time}}
\begin{center}
\resizebox{.8\textwidth}{!}{\begin{tabular}{cccccccccc}
\toprule
\multirow{2}{*}{Codes} & \multirow{2}{*}{Parameter} & \multicolumn{2}{c}{FLOPs} & \multicolumn{2}{c}{Inference (codeword)} & \multicolumn{2}{c}{Training (epoch)} & \multicolumn{2}{c}{Mask density}\\
\cmidrule(r){3-4}\cmidrule(r){5-6}\cmidrule(r){7-8}\cmidrule(r){9-10}
                                &                            & CrossMPT  & ECCT     & CrossMPT  & ECCT     & CrossMPT  & ECCT  & CrossMPT  & ECCT   \\ 
\midrule
\multirow{1}{*}{BCH}   & (63,45)                             & 11.8~M               & 14.0~M              & 326~$\mu$s                & 328~$\mu$s              & 29~s                 & 29~s              & 32.45\%               & 53.09\%  \\\midrule 
\multirow{2}{*}{LDPC}  & (121,70)                            & 28.8~M               & 37.7~M              & 400~$\mu$s                & 450~$\mu$s              & 58~s                 & 80~s              & 9.09\%                & 24.01\%  \\ 
                       & (121,80)                            & 26.3~M               & 34.6~M              & 391~$\mu$s                & 436~$\mu$s              & 53~s                 & 76~s              & 9.09\%                & 21.94\%  \\\midrule 
\multirow{1}{*}{Turbo} & (132,40)                            & 41.8~M               & 55.0~M              & 459~$\mu$s                & 511~$\mu$s              & 83~s                 & 110~s             & 11.43\%               & 14.25\% \\\midrule 
\multirow{1}{*}{BCH}  & (255,223)                           & 4.43~M               & 12.9~M              & 747~$\mu$s               & 859~$\mu$s             & 56~s                & 145~s             & 48.63\%                & 78.21\%  
\\\midrule 
\multirow{1}{*}{WRAN}  & (384,320)                           & 10.0~M               & 29.3~M              & 1295~$\mu$s               & 1638~$\mu$s             & 104~s                & 305~s             & 5.21\%                & 13.25\% \\ 
\bottomrule
\end{tabular}}
\end{center}
\end{table}

\section{Conclusion}

We developed a novel transformer architecture for ECC decoding called CrossMPT, which improves both decoding performance and computational complexity.
CrossMPT achieves this by adopting a more effective architecture that processes magnitude and syndrome through the cross-attention mechanism.
This approach leverages the clear and compact representation of codeword bit relationships in the PCM, enabling the model to accurately learn these relationships while also reducing memory usage, FLOPs, inference time, and training time.
Most existing research on transformer-based decoders has primarily focused on short codes due to challenges in training long codes, caused by high memory usage and complexity.
However, CrossMPT offers a potential breakthrough, paving the way for transformer-based decoders to be effectively applied to long codes.

\bibliography{iclr2025_conference}
\bibliographystyle{iclr2025_conference}

\appendix

\section{Performance on Longer Codes}
\label{append_long}
\begin{figure*}[!h]
\begin{center}
\subfigure[(529,440) LDPC code \label{fig_long_a}]{\includegraphics[width=.32\textwidth]{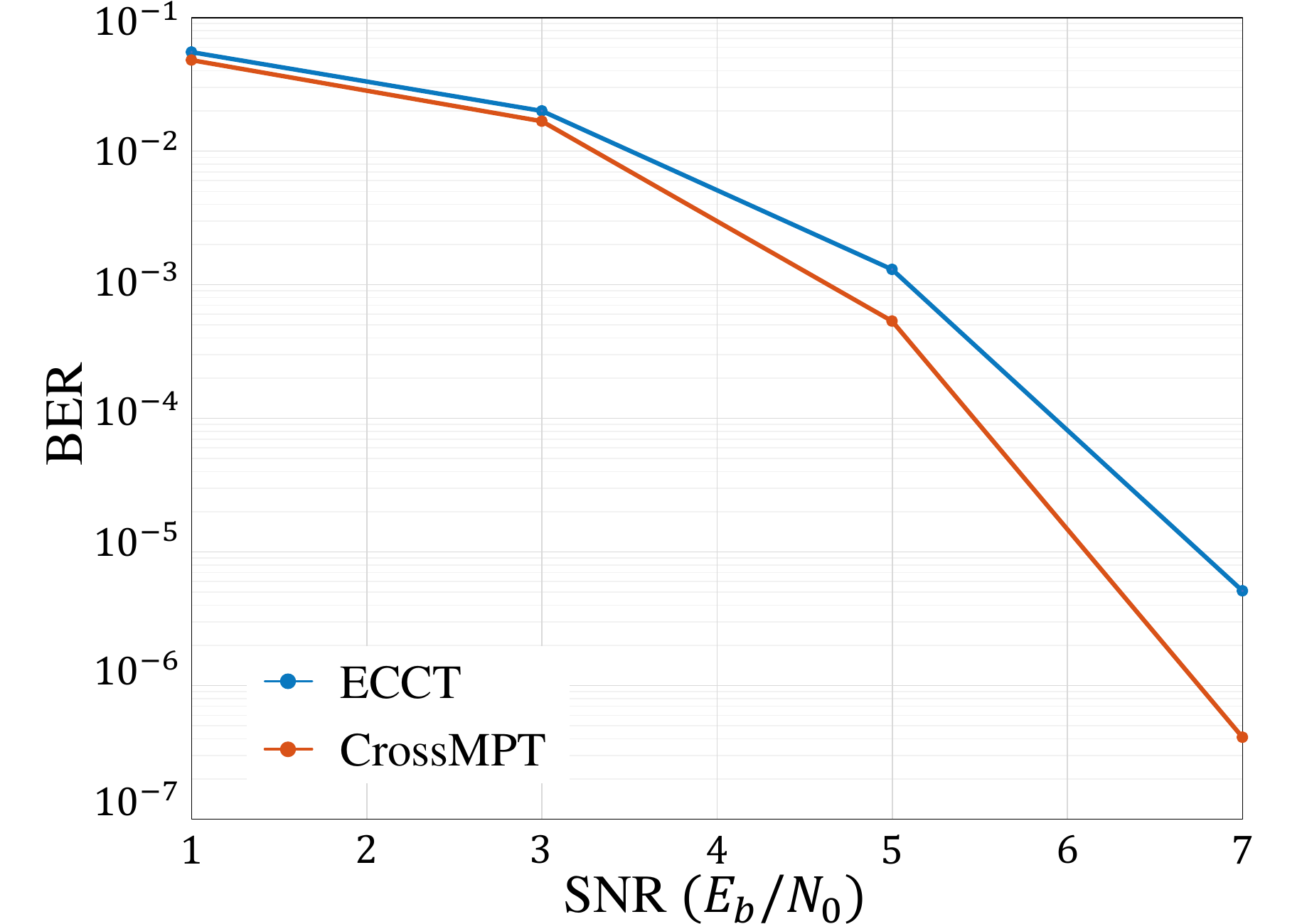}}
\subfigure[(384,320) WRAN LDPC code \label{fig_long_b}]{\includegraphics[width=.32\textwidth]{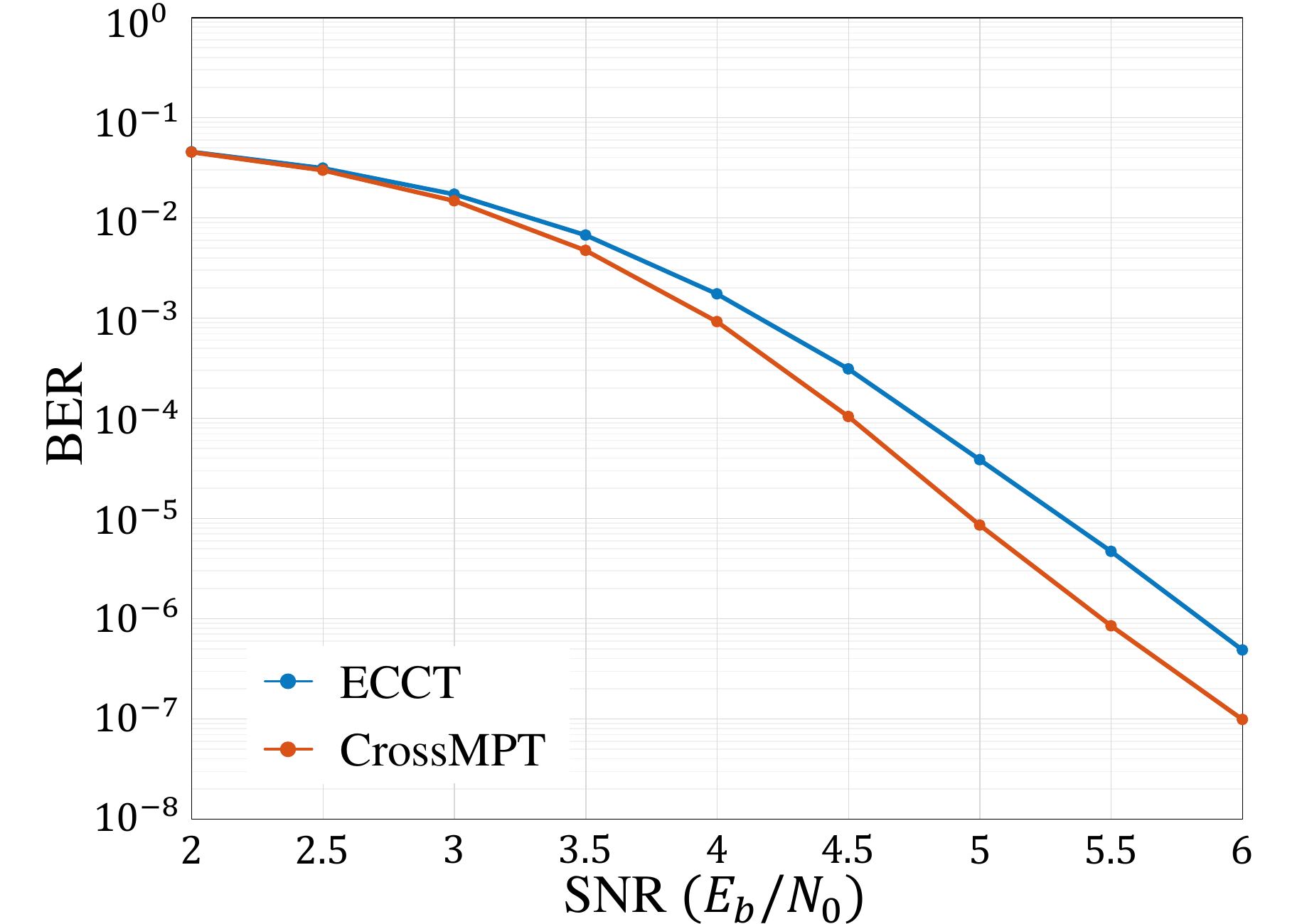}}
\subfigure[(512,384) polar code \label{fig_long_c}]{\includegraphics[width=.32\textwidth]{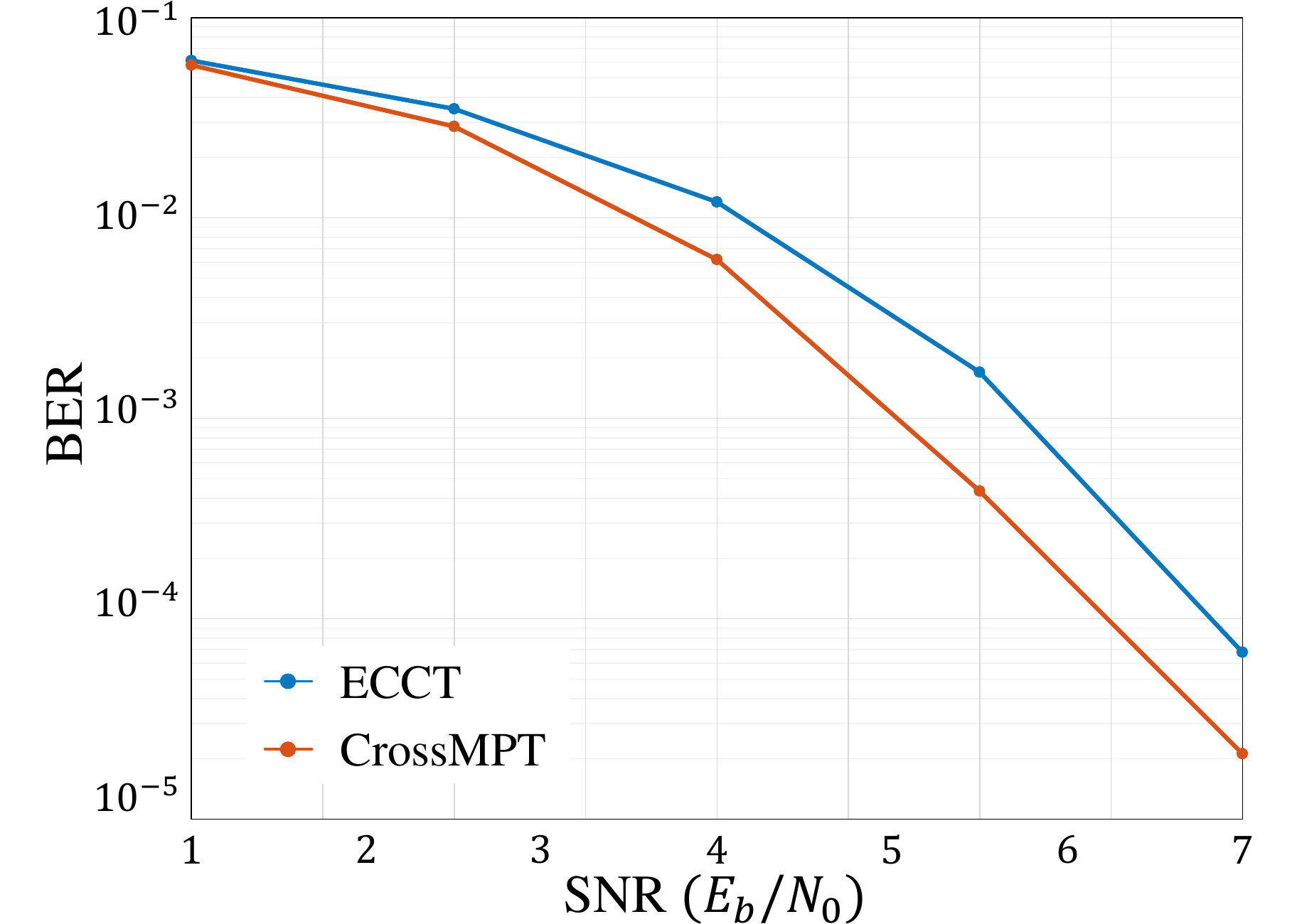}}
\vfill
\subfigure[(648,540) IEEE802.11n \label{fig_long_d}]{\includegraphics[width=.32\textwidth]{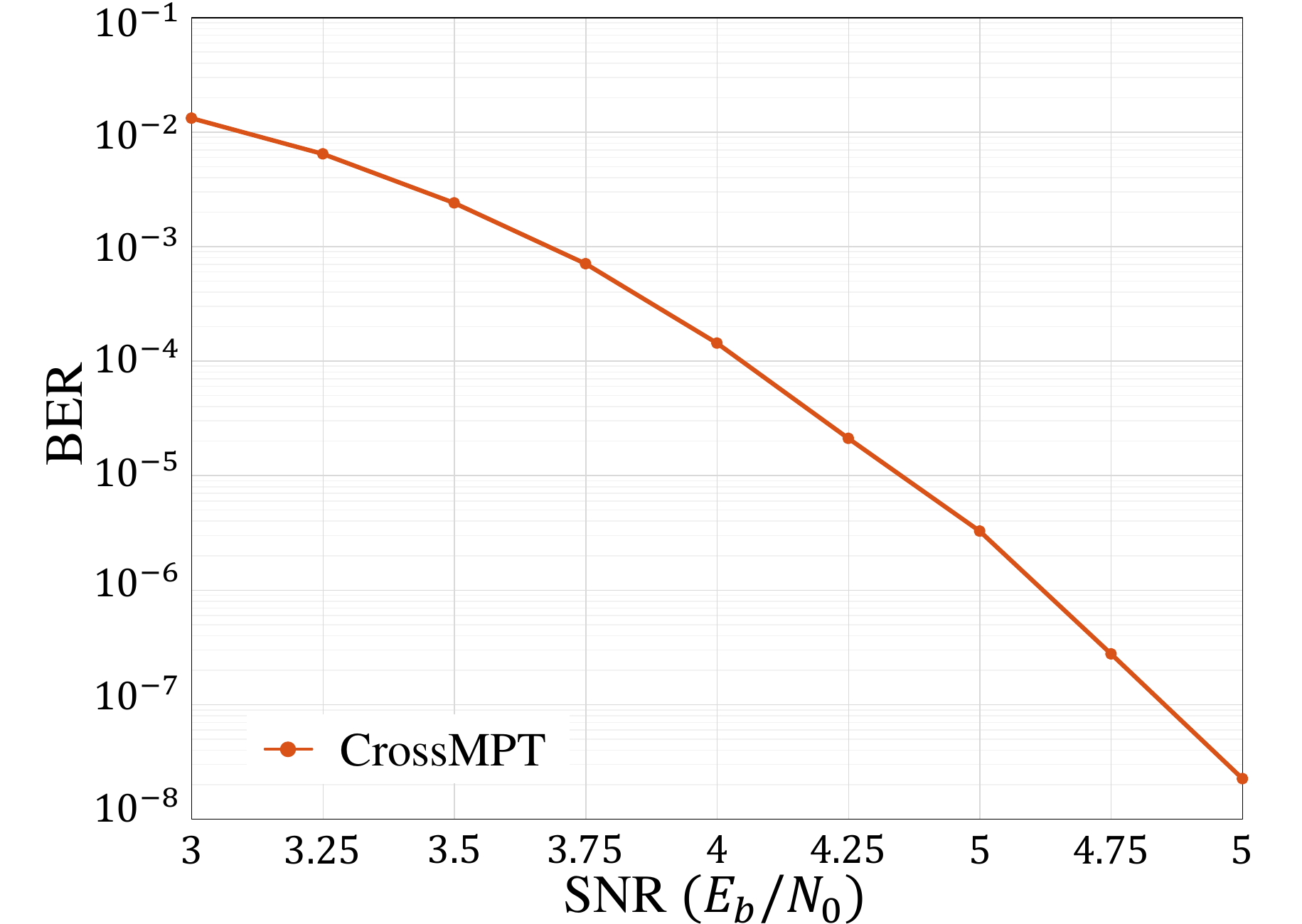}}
\subfigure[(1056,880) WiMAX \label{fig_long_e}]{\includegraphics[width=.32\textwidth]{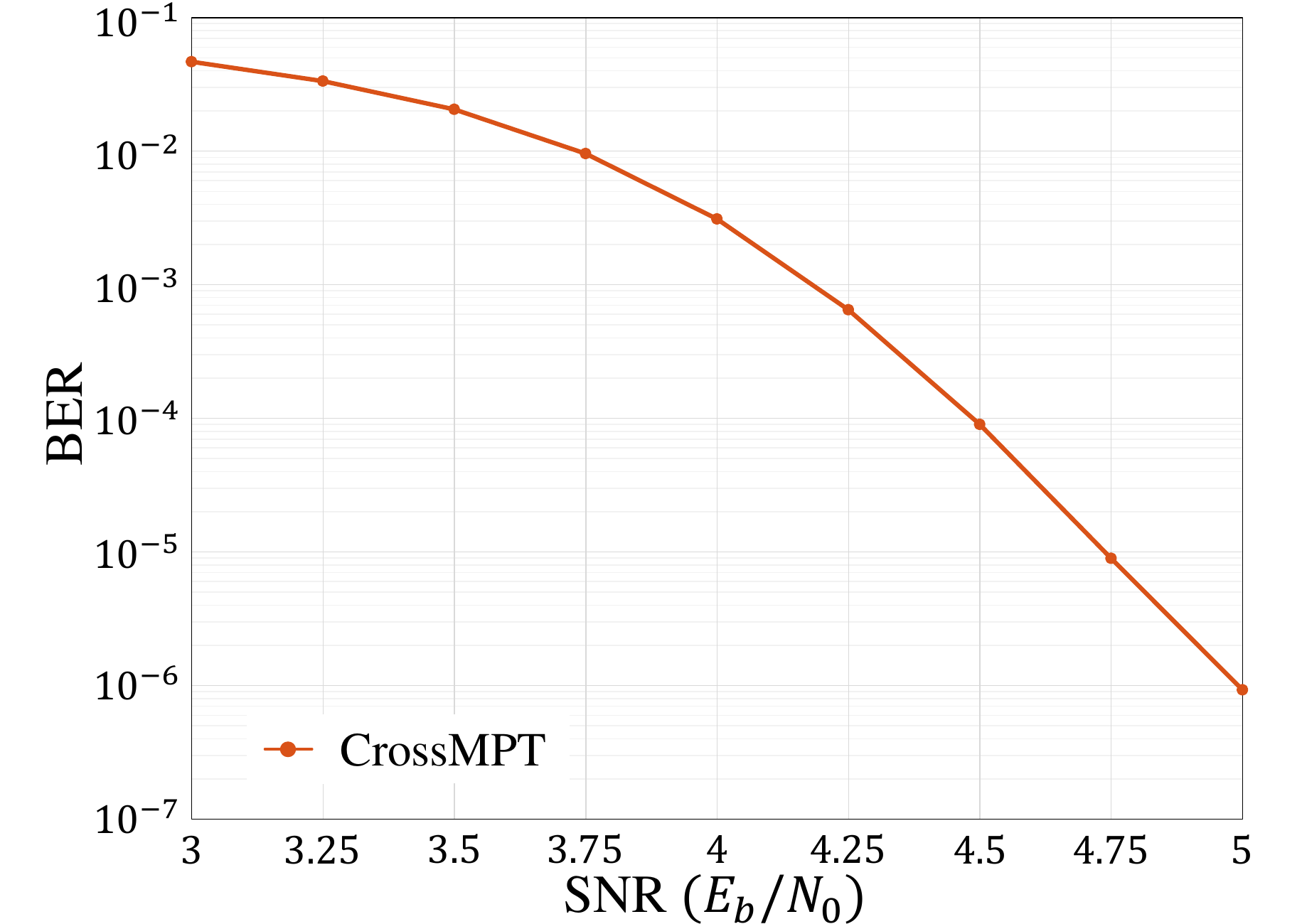}}
\caption{The decoding performance of long codes.
\label{fig_larger_graph1}}
\end{center}
\end{figure*}

We present the BER performance for three longer codes in Figures~\ref{fig_long_a}, \ref{fig_long_b}, and \ref{fig_long_c} for ECCT and CrossMPT $N=6$, $d=32$.
For all three codes ((a) (529,440) LDPC code, (b) (384,320) wireless regional area network~(WRAN) LDPC code, (c) (512,384) polar code), the proposed CrossMPT outperforms the original ECCT.
Despite its reduced complexity, CrossMPT significantly enhances the decoding performance compared to ECCT, not only for short-length codes but also for longer codes.
Also, Figures~\ref{fig_long_d} and \ref{fig_long_e} show the decoding performance of CrossMPT for much longer codes.
The BER performances of the (648,540) IEEE802.11n LDPC code~($N=10$, $d=128$) and (1056,880) WiMAX LDPC code~($N=6$, $d=32$) demonstrate that CrossMPT efficiently trains how to decode the codeword even for large $N$ and $d$ and performs well for longer codes.
Again, we emphasize CrossMPT's capability to decode long codes where ECCT struggles due to high memory allocation (large attention map).
The structure of CrossMPT demonstrates its efficiency in learning long codes, surpassing the limitations of short or moderate codelengths of transformer-based decoders.

\section{Comparison with the BP decoder}
\label{append_BP}
\begin{figure*}[!h]
\begin{center}
\subfigure[(121,80) LDPC code\label{fig_larger_graph_a}]{\includegraphics[width=.45\textwidth]{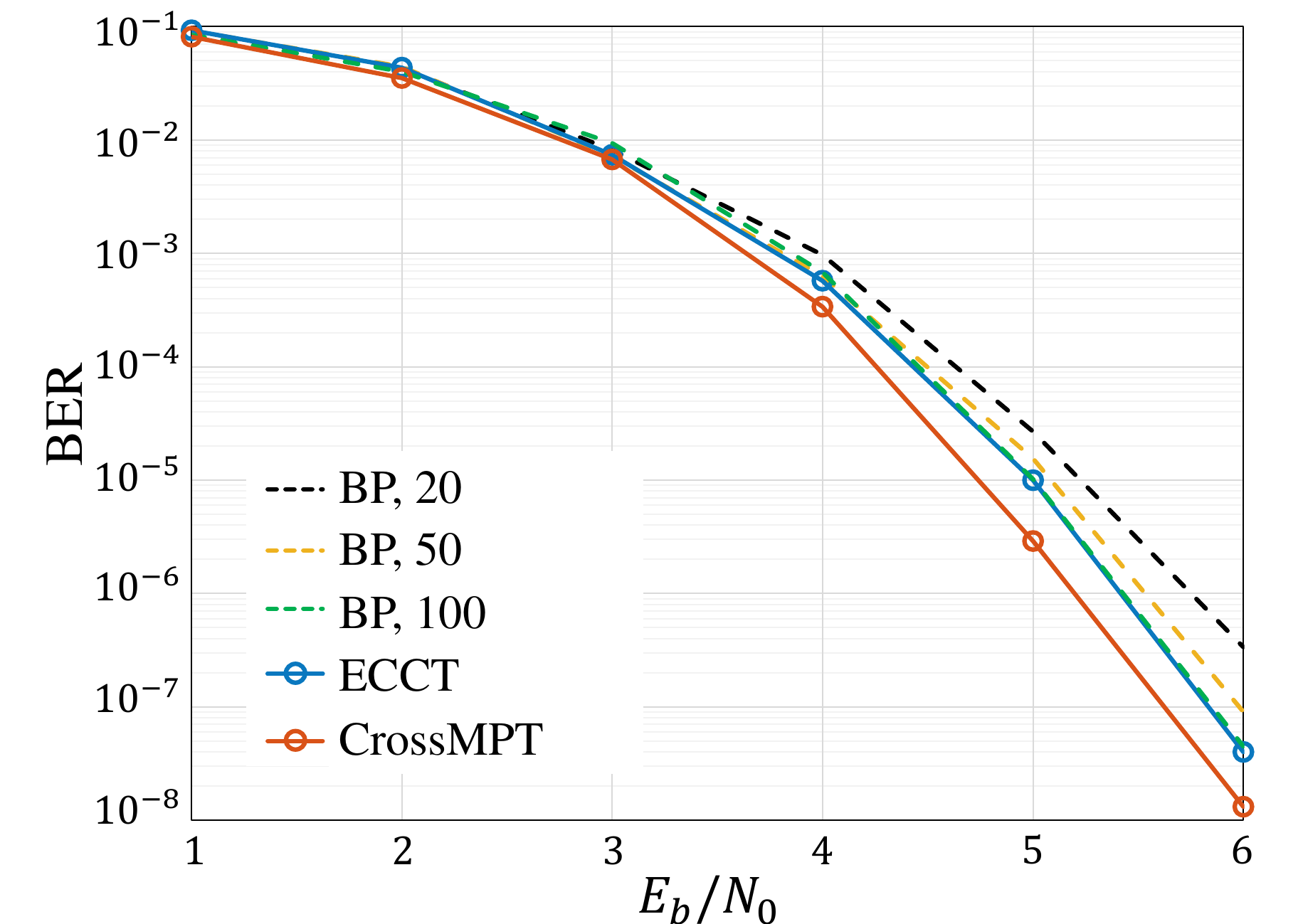}}
\hfill
\subfigure[(648,540) IEEE 802.11n\label{fig_larger_graph_b}]{\includegraphics[width=.45\textwidth]{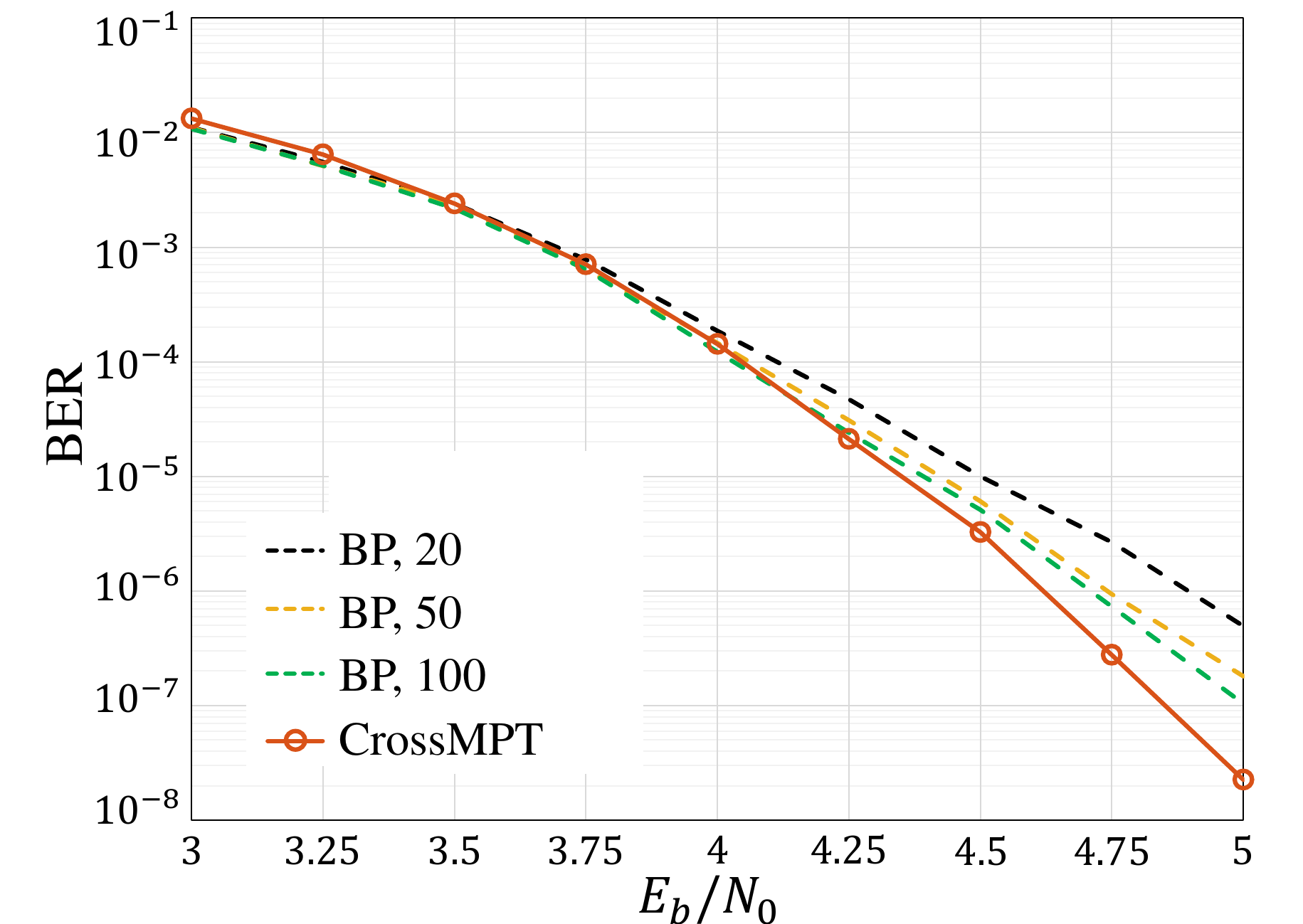}}
\caption{Performance comparison between BP decoder~(iteration 20, 50, and 100) and CrossMPT.
\label{fig_larger_BP}}
\end{center}
\end{figure*}

Figure~\ref{fig_larger_BP} shows the decoding performance between the traditional BP decoder with a maximum number of iterations of 20, 50, 100, and 200 and CrossMPT for both
short and long LDPC codes.
Figures~\ref{fig_larger_graph_a} and \ref{fig_larger_graph_b} compare the BER performance for (121,80) LDPC codes~($N=6$, $d=128$) and (648,540) IEEE 802.11n LDPC code~($N=10$, $d=128$), respectively.
Notably, the proposed CrossMPT can outperform the BP decoder for both short and long LDPC codes.
These results highlight that CrossMPT efficiently trains how to decode the codeword across a wide range of code lengths.

\section{Comparison with the ML decoder}
\label{append_ML}
We compare ECCT and CrossMPT with the ML decoder for short BCH codes.
Figure~\ref{fig_larger_BP} demonstrates the BER performance of $(31,16)$ BCH code and $(31,21)$ BCH code.
Especially, these results show that CrossMPT closely approaches the optimal ML performance for short codes.

\begin{figure*}[!h]
\begin{center}
\subfigure[(31,16) BCH code\label{fig_ML_a}]{\includegraphics[width=.45\textwidth]{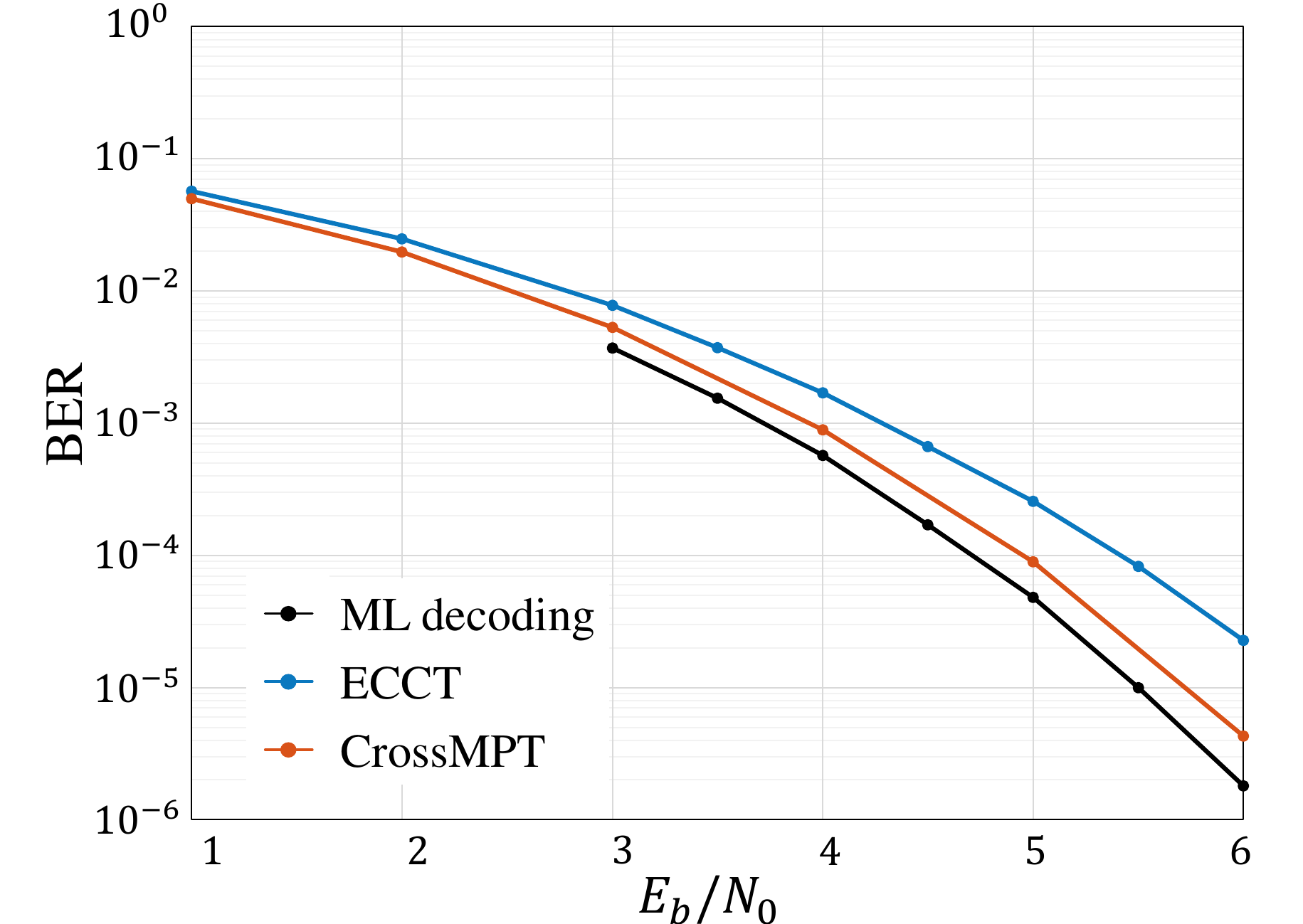}}
\hfill
\subfigure[(31,21) BCH code\label{fig_ML_b}]{\includegraphics[width=.45\textwidth]{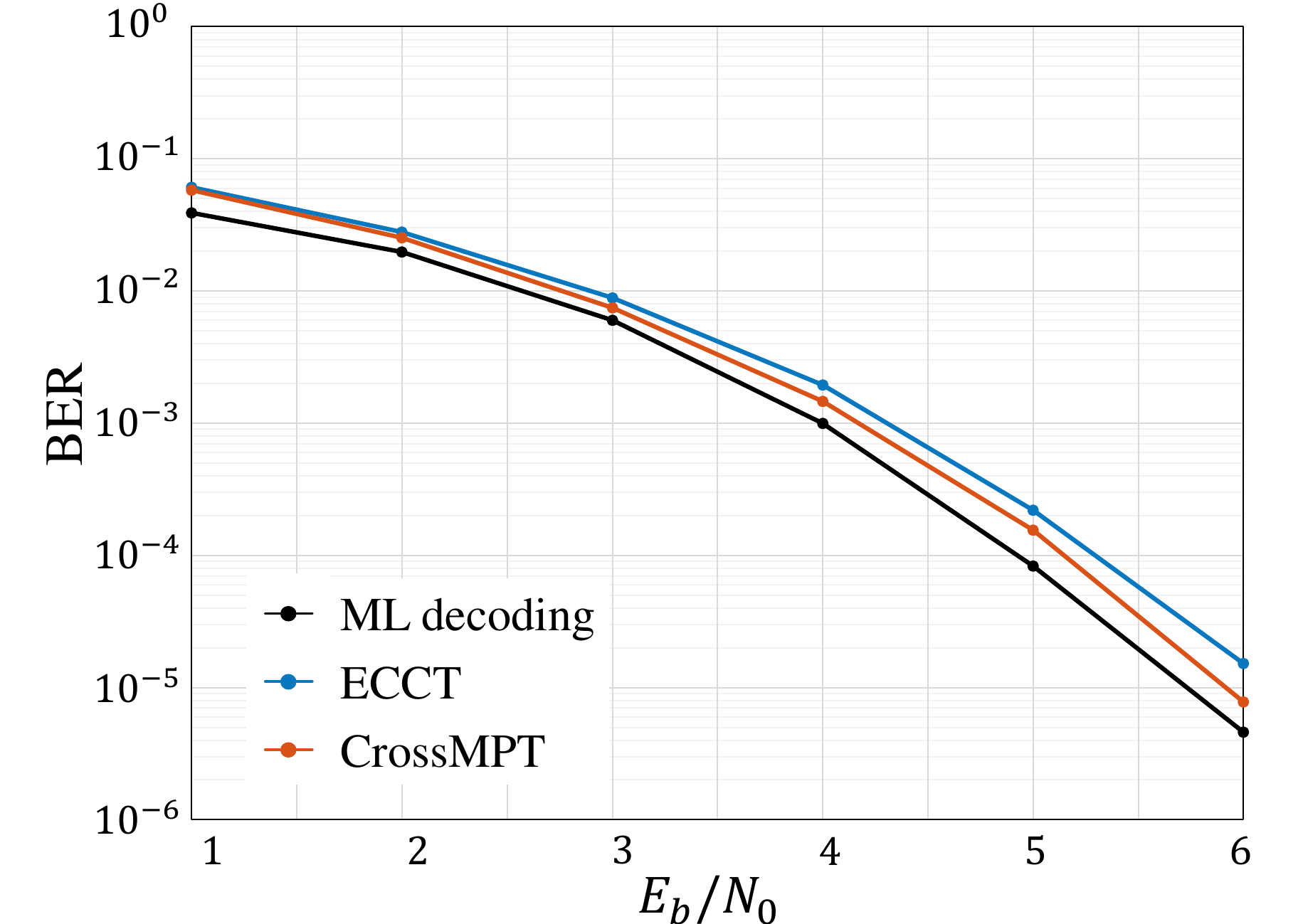}}
\caption{The decoding performance comparison between ML decoder, ECCT, and CrossMPT.
\label{fig_ML}}
\end{center}
\end{figure*}

\section{Comparison with Successive Cancellation List Polar Decoder}
\label{append_SCL}
We compare the BER performance of the SCL decoder, ECCT, and CrossMPT in Table~\ref{tab_appendix_D}.
The performance of the SCL decoder is from~\citep{b_ECCT}.
Although the contribution of $L$ is significant in long codes, the SCL decoder achieves a great performance with small $L$, such as $L=4$.
As reported in~\citep{b_ECCT,b_DDECC}, the SCL decoder outperforms ECCT.
This is because the SCL decoder is a decoder specialized for Polar codes and is a state-of-the-art algorithm that has undergone extensive development over a long period.
CrossMPT has made significant improvements from ECCT and even outperforms the SCL decoder for (64,48) polar code.

\begin{table}[!h]
\caption{Comparison of decoding performance at three different SNR values (4 dB, 5 dB, 6 dB) for SCL decoder, ECCT, and  CrossMPT. The results are measured by the negative natural logarithm of BER.
The best results are highlighted in \textbf{bold} and the second best is \underline{underlined}.
Higher is better.\label{tab_appendix_D}}
\begin{center}
\resizebox{\textwidth}{!}{\begin{tabular}{ccccccccccccc}
\toprule
\multicolumn{1}{c}{Method} & \multicolumn{3}{c}{SCL ($L=1$)} & \multicolumn{3}{c}{SCL ($L=4$)} & \multicolumn{3}{c}{ECCT} & \multicolumn{3}{c}{CrossMPT} \\
\cmidrule(r){1-1}\cmidrule(r){2-4}\cmidrule(r){2-4}\cmidrule(r){5-7}\cmidrule(r){8-10}\cmidrule(r){11-13}
\multicolumn{1}{c}{Parameter}    & 4   & 5    & 6    & 4   & 5    & 6    & 4    & 5    & 6    & 4     & 5      & 6     \\
\midrule
(64,32)                    & 7.30  & 9.67   & 13.18  & \textbf{8.11}  & \textbf{10.70}  & \textbf{14.04}  & 6.99   & 9.44   & 12.32  & \underline{7.50}    & \underline{9.97}     & \underline{13.31}   \\
(64,48)                    & 6.19  & 8.41   & 10.97  & \textbf{6.69}  & 8.63   & 11.24  & 6.36   & 8.46   & 11.09  & 6.51    & \textbf{8.70}     & \textbf{11.31}   \\
(128,64)                   & 8.37  & 11.69  & 13.70  & \textbf{9.60}  & \textbf{13.16}  & \textbf{17.42}  & 5.92   & 8.64   & 12.18  & \underline{7.52}    & \underline{11.21}    & \underline{14.76}   \\
(128,86)                   & 7.54  & 10.74  & 15.14  & \textbf{9.26}  & \textbf{13.04}  & \textbf{17.13}  & 6.31   & 9.01   & 12.45  & \underline{7.86}    & \underline{11.45}    & \underline{15.47}   \\
(128,96)                   & 6.74  & 9.53   & 13.53  & \textbf{8.02}  & \textbf{11.60}  & \textbf{18.16}  & 6.31   & 9.12   & 12.47  & \underline{7.15}    & \underline{10.15}    & \underline{13.13}\\
\bottomrule
\end{tabular}}
\end{center}
\end{table}

\section{Comparison with DDECCT}
\label{append_DDECCT}
For a fair comparison with DDECCT, we also apply the denoising diffusion training technique to CrossMPT.
Table~\ref{tab_appendix_B} compares the BER performance of ECCT~\citep{b_ECCT}, CrossMPT, DDECCT, and CrossMPT applying the denoising diffusion model.
All four decoders are model-free decoders using the transformer architecture, and simulations are taken for $N=6$, $d=128$.
We conduct simulations for codes where DDECC performs better than CrossMPT.
For the rest of the codes, CrossMPT outperforms DDECC.
The proposed CrossMPT shows superior decoding performance compared to the original ECCT.
Compared to DDECC, CrossMPT demonstrates similar BER performance for polar codes, but it even outperforms DDECC for BCH and LDPC codes.
When the denoising diffusion technique is applied to CrossMPT, it achieves the best performance among others, where DDECC, CrossMPT, and ECCT follow.
This proves that the CrossMPT architecture provides separate gain from the denoising diffusion algorithm for transformer-based decoders.

\begin{table*}[!h]
\caption{Comparison of decoding performance at three different SNR values (4 dB, 5 dB, 6 dB) for ECCT~\citep{b_ECCT}, CrossMPT, and DDECC~\citep{b_DDECC}. The results are measured by the negative natural logarithm of BER. The best results are highlighted in \textbf{bold} and the second best is \underline{underlined}. Higher is better.\label{tab_appendix_B}}
\begin{center}
\resizebox{\textwidth}{!}{\begin{tabular}{cccccccccccccc}
\toprule
\multicolumn{2}{c}{Architecture}                 & \multicolumn{6}{c}{\textit{Without} denoising diffusion}                                                                  & \multicolumn{6}{c}{\textit{With} denoising diffusion}      \\
\cmidrule(r){1-2}\cmidrule(r){3-8}\cmidrule(r){9-14}
\multirow{2}{*}{Codes} & \multirow{2}{*}{Parameter} & \multicolumn{3}{c}{ECCT}                        & \multicolumn{3}{c}{CrossMPT}                    & \multicolumn{3}{c}{ECCT}                        & \multicolumn{3}{c}{CrossMPT}                       \\
\cmidrule(r){3-5}\cmidrule(r){6-8}\cmidrule(r){9-11}\cmidrule(r){12-14}
                       &                         & 4             & 5              & 6              & 4             & 5              & 6              & 4             & 5              & 6     & 4             & 5              & 6         \\
                       \midrule
\multirow{1}{*}{BCH}   & (63,36)                 & {4.86} & {6.65}  & {9.10}  & 5.03          & 6.91           & 9.37           & \underline{{5.11}} & \underline{{7.09}}  & \underline{{9.82}} & \textbf{5.23}    & \textbf{7.20}   & \textbf{10.01} \\
                       \midrule
\multirow{3}{*}{Polar} & (128,64)                & 5.92          & {8.64}  & {12.18} & 7.52          & 11.21          & 14.76          & \underline{{9.11}} & \underline{{12.9}}  & \underline{{16.30}}  & \textbf{10.21}   & \textbf{13.63}  & \textbf{17.28}\\
                       & (128,86)                & 6.31          & 9.01           & 12.45          & 7.51          & \underline{{10.83}} & \underline{{15.24}} & \underline{{7.60}}  & {10.81} & {15.17} & \textbf{8.56}    & \textbf{12.04}  & \textbf{15.37}\\
                       & (128,96)                & {6.31} & {9.12}  & {12.47} & 7.15          & 10.15          & 13.13          & \underline{{7.16}} & \underline{{10.3}}  & \underline{{13.19}} & \textbf{7.57}    & \textbf{10.61}  & \textbf{13.33}\\
                       \midrule
MacKay                 & (96,48)                 & {7.38} & 10.72          & 14.83          & 7.97          & 11.77          & 15.52          & \underline{{8.12}} & \underline{{11.88}} & \underline{{15.93}} & \textbf{8.85}    & \textbf{12.58}  & \textbf{17.69}\\
\bottomrule
\end{tabular}}
\end{center}
\end{table*}

\section{Decoding Performance for Rayleigh Fading Channel}
\label{append_Rayleigh}
The original ECCT architecture shows robustness to non-Gaussian channels (e.g., Rayleigh fading channel)~\citep[Supplementary]{b_ECCT}.
We also measured the decoding performance of CrossMPT in Rayleigh fading channels. To compare with ECCT, we use the same fading channel as in~\citep{b_ECCT}. The received codeword is given as $y=hx+z$, where $h$ is an $n$-dimensional i.i.d. Rayleigh distributed vector with a scale parameter $\alpha=1$ and $z\sim N(0,\sigma^2)$. The following table demonstrates the BER performance of ECCT and CrossMPT in Rayleigh fading channel and CrossMPT still outperforms the original ECCT architecture for all types of codes.

\begin{table}[!h]
\resizebox{\textwidth}{!}{\begin{tabular}{ccccccccccccccccccc}
\toprule
Codes    & \multicolumn{3}{c}{(31,16) BCH} & \multicolumn{3}{c}{(64,32) Polar} & \multicolumn{3}{c}{(128,64) Polar} & \multicolumn{3}{c}{(128,86) Polar} & \multicolumn{3}{c}{(121,70) LDPC} & \multicolumn{3}{c}{(128,64) CCSDS} \\
\cmidrule(r){1-1}\cmidrule(r){2-4}\cmidrule(r){5-7}\cmidrule(r){8-10}\cmidrule(r){11-13}\cmidrule(r){14-16}\cmidrule(r){17-19}
Methods      & 4         & 5        & 6        & 4         & 5         & 6         & 4          & 5         & 6         & 4          & 5         & 6         & 4         & 5         & 6         & 4          & 5         & 6         \\
\midrule
ECCT     & 5.18      & 6.04     & 6.92     & 5.53      & 6.62      & 7.80      & 4.31       & 5.37      & 6.63      & 4.02       & 4.81      & 5.70      & 3.91      & 4.97      & 6.31      & 2.46       & 3.97      & 5.79      \\
CrossMPT & 5.53      & 6.55     & 7.61     & 5.91      & 7.17      & 8.48      & 4.70       & 5.93      & 7.34      & 4.41       & 5.38      & 6.46      & 4.25      & 5.53      & 7.11      & 5.25       & 6.94      & 8.92     \\
\bottomrule

\end{tabular}}
\end{table}

\section{Ablation Study with Additional Masking}
\label{append_ablation_masking}

To understand the impact of magnitude-magnitude and syndrome-syndrome relationships, we demonstrate the decoding performance of ECCT with additional masking of these relationships.
Table~\ref{tab_appendix_ablation_mask} compares the decoding performance of ECCT with this additional masking, standard ECCT, and CrossMPT.
The results show no significant performance degradation with the additional masking, indicating that the magnitude-magnitude and syndrome-syndrome relationships are not critical to decoding performance.

\begin{table}[!h]
\caption{Comparison of decoding performance at three different SNR values (4 dB, 5 dB, 6 dB) for ECCT with this additional masking, standard ECCT, and CrossMPT. The results are measured by the negative natural logarithm of BER.
The best results are highlighted in \textbf{bold}.
Higher is better.\label{tab_appendix_ablation_mask}}
\begin{center}
\resizebox{.75\textwidth}{!}{\begin{tabular}{cccccccccc}
\toprule
\multicolumn{1}{c}{Method} & \multicolumn{3}{c}{ECCT + Masking} & \multicolumn{3}{c}{ECCT} & \multicolumn{3}{c}{CrossMPT}\\
\cmidrule(r){1-1}\cmidrule(r){2-4}\cmidrule(r){5-7}\cmidrule(r){8-10}
\multicolumn{1}{c}{Parameter}    & 4   & 5    & 6    & 4   & 5    & 6    & 4   & 5    & 6\\
\midrule
$(31,16)$ BCH   & 6.52 & 8.55 & 11.42 & 6.39 & 8.29 & 10.66 & \textbf{6.98} & \textbf{9.25} & \textbf{12.48} \\
$(63,45)$ BCH   & 5.53 & 7.74 & 10.88 & 5.60 & 7.79 & 10.93 & \textbf{5.90} & \textbf{8.20} & \textbf{11.62} \\
$(64,48)$ Polar   & 6.25 & 8.26 & 10.93 & 6.36 & 8.46 & 11.09 & \textbf{6.51} & \textbf{8.70} & \textbf{11.31} \\
$(121,60)$ LDPC  & 4.98 & 7.91 & 12.61 & 5.17 & 8.31 & 13.30 & \textbf{5.74} & \textbf{9.26} & \textbf{14.78} \\
\bottomrule
\end{tabular}}
\end{center}
\end{table}

\section{Visualization of Cross-attention Map}
\label{append_attn_avg}

\begin{figure}[!h]
\begin{center}
\centerline{\includegraphics[width=.75\columnwidth]{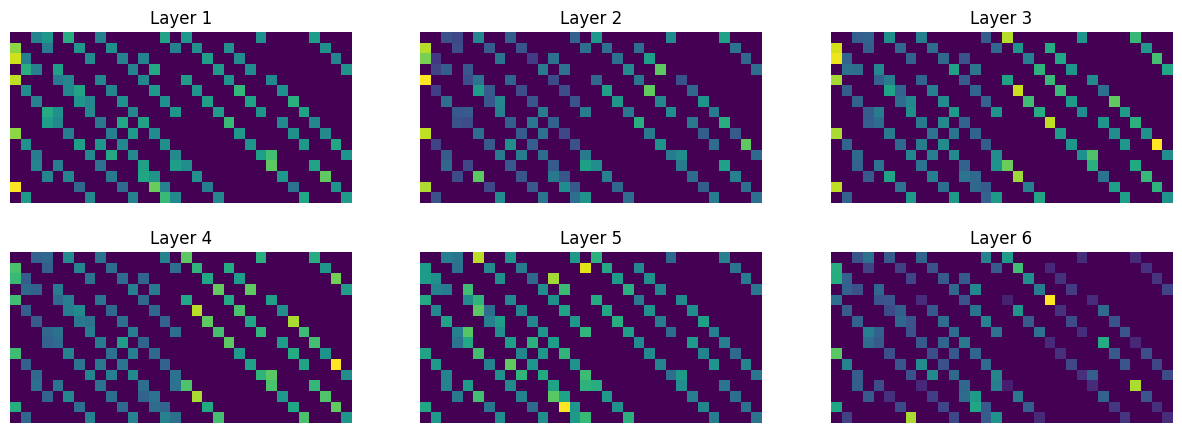}}
\caption{Attention scores of $N=6$ layers with a single bit error.\label{fig_append_attn_avg}}
\end{center}
\end{figure}

Figure~\ref{fig_append_attn_avg} illustrates the attention scores for all $N=6$ layers with a single bit error~(bit error in the \textit{first position}).
The first three layers have relatively high attention score at the error position~(first bit).
Then, when the error is corrected, the attention score becomes lower at the last three layers.

\section{Impact of the Proposed Mask Matrix and Training Convergence}
\label{append_conv}
\begin{figure*}[!h]
\begin{center}
\subfigure[$(128,86)$ polar code]{\includegraphics[width=.45\textwidth]{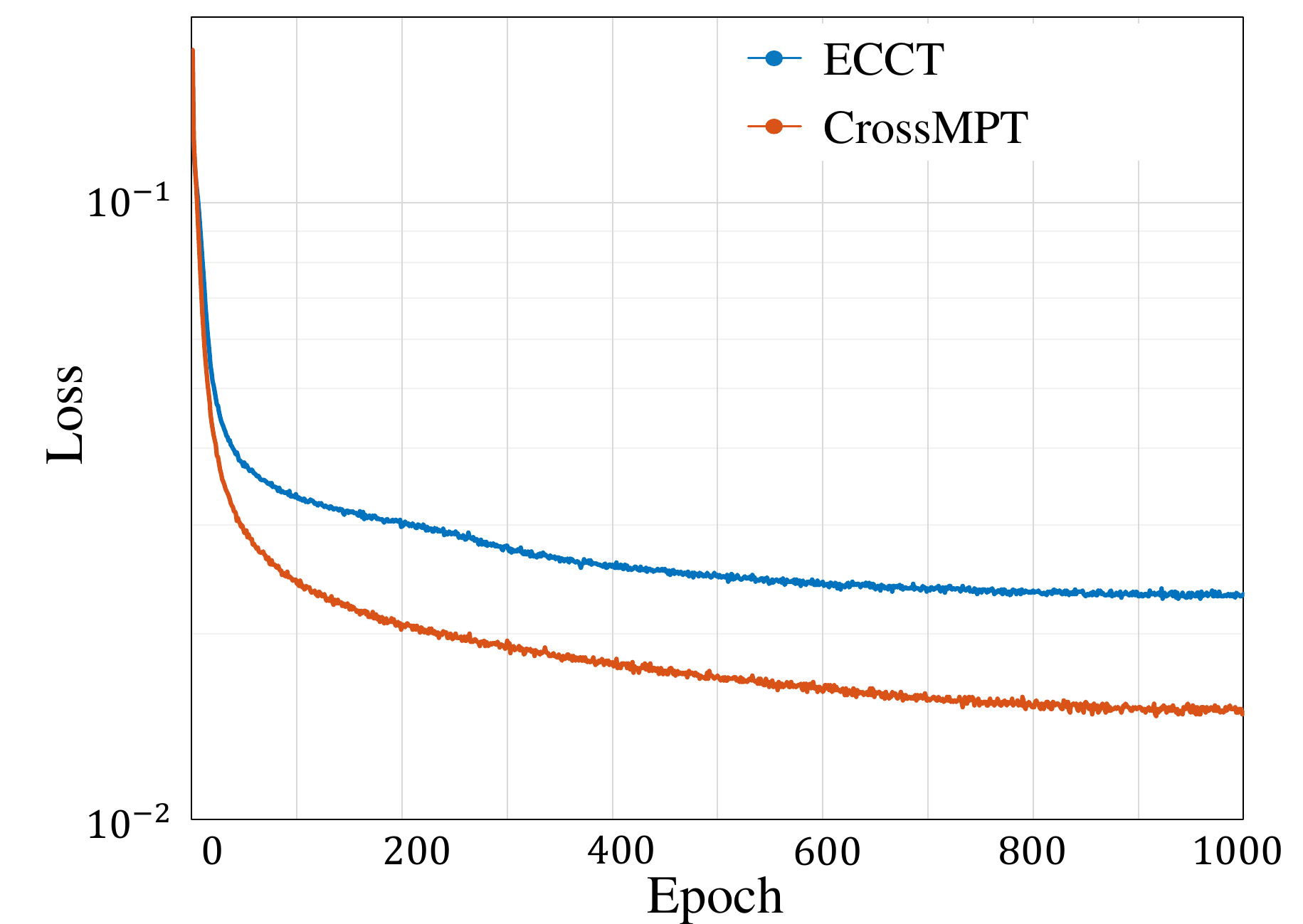}}
\hfill
\subfigure[$(121,70)$ LDPC code]{\includegraphics[width=.45\textwidth]{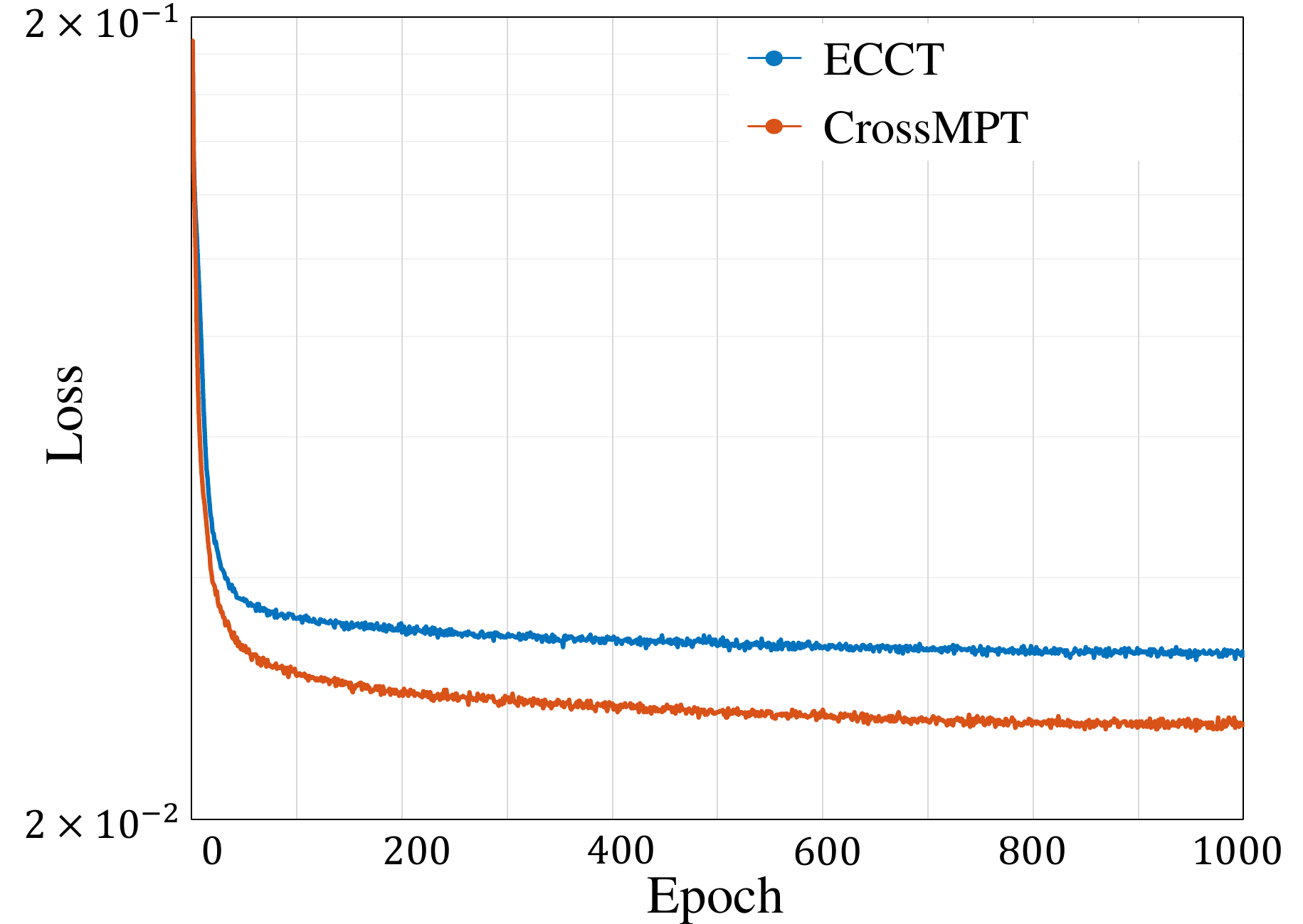}}
\caption{Comparison of the training loss of ECCT and CrossMPT.
\label{fig_conv}}
\end{center}
\end{figure*}

The mask matrix in transformer-based decoders enables the model to efficiently learn the relevance between input bits.
In the masked cross-attention module in CrossMPT, the mask matrix is the PCM itself, utilizing only the essential information of codeword bits.
To demonstrate the impact of the CrossMPT architecture, we observe the behavior of the loss at each epoch.
Fig.~\ref{fig_conv} compares the training convergence between ECCT and CrossMPT.
The training of CrossMPT is much faster than ECCT, demonstrating CrossMPT's efficiency in achieving superior decoding performance with limited training epochs.

\section{Throughput Analysis}

\begin{figure}[!h]
\begin{center}
\centerline{\includegraphics[width=.7\columnwidth]{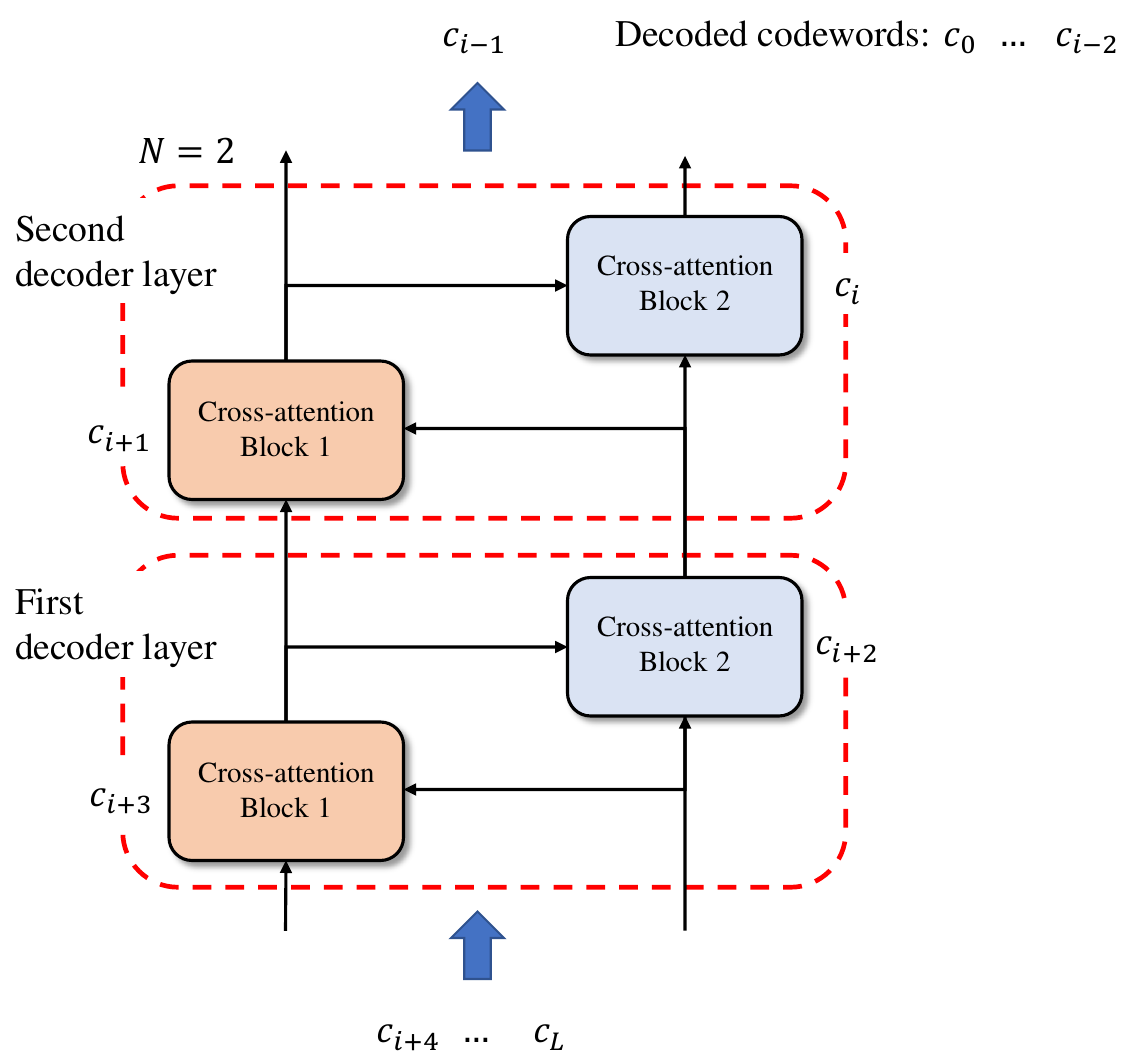}}
\caption{Example of decoding multiple codewords in CrossMPT with $N=2$.\label{fig_throughput}}
\end{center}
\end{figure}

If the decoder is implemented with specific hardware, pipelining approach can be used to maximize the throughput.
Employing pipelining, CrossMPT achieves a decoding throughput comparable to that of ECCT.
By unrolling $N$ layers and adapting the decoder for a fully parallel hardware architecture, CrossMPT can process two consecutive codewords simultaneously across two cross-attention blocks within the same layer.
This means that while the second cross-attention block is processing the first codeword, the first cross-attention block can concurrently decode the subsequent codeword. This pipelining strategy ensures that throughput levels remain comparable to that of ECCT.
Figure~\ref{fig_throughput} illustrates the example of decoding multiple codewords in CrossMPT with $N=2$:

In wireless communications, the decoder’s throughput is often a more critical concern than latency.
This is because the latency from communication protocols and signal processing in preceding receiver blocks would be longer than the latency introduced by the channel decoder.
Throughput becomes especially important when supporting very high data rates in wireless communication as the channel decoder can be a bottleneck.

Furthermore, in wireless communication scenarios, a sequential algorithm may be preferred for its enhanced performance or reduced complexity.
The layered decoding algorithm for LDPC codes has been widely adopted as a de facto standard~\citep{b_Bae2019, b_Hailes2015, b_Li2021}, despite its sequential nature and limitation on parallelism, exemplifying the preference for sequential algorithms.
The layered decoding algorithm is favored for its superior decoding performance compared to fully parallel sum-product decoding at equivalent computational complexities.

\label{append_throughput}

\end{document}